\documentclass[runningheads]{llncs}

\pdfoutput=1

\usepackage{graphicx}

\usepackage{tikz}
\usepackage{comment}
\usepackage{amsmath,amssymb} %
\usepackage{color}

\usepackage[accsupp]{axessibility}  %

\usepackage{graphicx}
\usepackage{amsmath}
\usepackage{amssymb}
\usepackage{booktabs}
\usepackage{multicol}
\usepackage{multirow}
\usepackage[para,online,flushleft]{threeparttable}
\usepackage{xcolor}
\usepackage{bm}
\usepackage{bbm}
\usepackage{cite}
\usepackage{algorithm}
\usepackage[noend]{algpseudocode}
\usepackage{stfloats}
\usepackage[pagebackref,breaklinks,citecolor=blue,colorlinks]{hyperref}
\usepackage{xr}

\usepackage[capitalize]{cleveref}
\crefname{section}{Sec.}{Secs.}
\Crefname{section}{Section}{Sections}
\Crefname{table}{Table}{Tables}
\crefname{table}{Tab.}{Tabs.}

\newcommand{\myparagraph}[1]{\vspace{0.2em}\noindent\textbf{#1}}

\begin{document}
\pagestyle{headings}
\mainmatter
\def\ECCVSubNumber{5754}  %

\title{SAGA: Stochastic Whole-Body \\ Grasping With Contact} %

\titlerunning{SAGA}
\author{Yan Wu$^*$\inst{1},
        Jiahao Wang$^*$\inst{2},
        Yan Zhang\inst{1},
        Siwei Zhang\inst{1},
        Otmar Hilliges\inst{1},
        Fisher Yu\inst{1},
        Siyu Tang\inst{1}}
\authorrunning{Y. Wu$^*$, J. Wang$^*$ et al.}

\institute{ETH Z\"urich, Switzerland \and Max Planck Institute for Informatics, Germany\\
\email{yan.wu@vision.ee.ethz.ch, jiwang@mpi-inf.mpg.de, \{yan.zhang,siwei.zhang,otmar.hilliges,siyu.tang\}@inf.ethz.ch, i@yf.io}}
\maketitle

\def\thefootnote{*}\footnotetext{Equal contribution.}

\begin{abstract}
The synthesis of human grasping has numerous applications including AR/VR, video games and robotics.
While methods have been proposed to generate realistic hand--object interaction for object grasping and manipulation, these typically only consider interacting hand alone.
Our goal is to \textbf{synthesize whole-body grasping motions}. Starting from an arbitrary initial pose, we aim to generate diverse and natural whole-body human motions to approach and grasp a target object in 3D space.
This task is challenging as it requires modeling both whole-body dynamics and dexterous finger movements. To this end, we propose \textbf{SAGA} (StochAstic whole-body Grasping with contAct), a framework which consists of two key components: %
(a) Static whole-body grasping pose generation. Specifically, we propose a multi-task generative model, to jointly learn static whole-body grasping poses and human-object contacts. 
(b) Grasping motion infilling. Given an initial pose and the generated whole-body grasping pose as the start and end of the motion respectively, we design a novel contact-aware generative motion infilling module to generate a diverse set of grasp-oriented motions. We demonstrate the effectiveness of our method, which is a novel generative framework to synthesize realistic and expressive whole-body motions that approach and grasp randomly placed unseen objects. Code and models are available at \href{https://jiahaoplus.github.io/SAGA/saga.html}{https://jiahaoplus.github.io/SAGA/saga.html}.

\end{abstract}

\section{Introduction}
\label{sec:intro}

\begin{figure*}
\begin{center}
    \centering
    \includegraphics[width=1\linewidth]{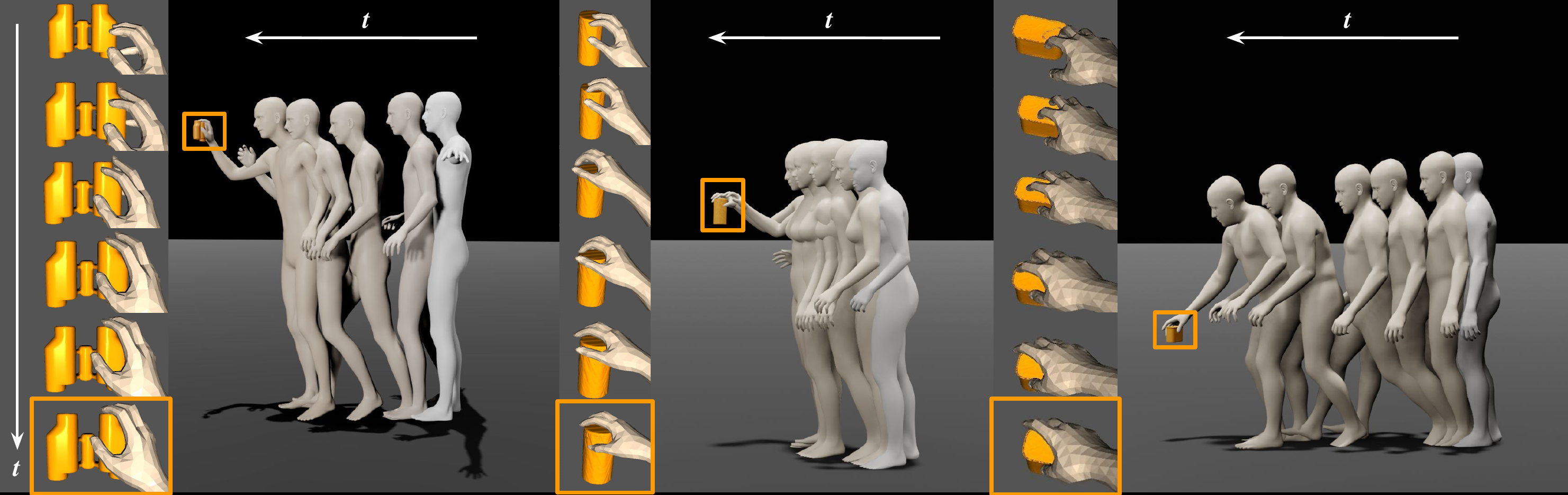}
    \caption{Generated whole-body grasping motion sequences (in beige) starting from a given pose (in white) to approach and grasp randomly placed unseen objects. For each sample, we present hand motion details in the last few frames on the left column. %
    }
    \label{fig:teaser}
\end{center}
\vspace{-1em}
\end{figure*}

A fully automated system that synthesizes realistic 3D human bodies approaching and grasping a target object in 3D space will be valuable in various fields, from robotics and animation to computer vision. Although remarkable progress has been made towards synthesizing realistic hand--object interactions, most existing works only focus on hand pose synthesis without considering whole-body movements~\cite{GRAB:2020, GrapingField:3DV:2020, jiang2021graspTTA}. Meanwhile, whole-body motion synthesis~\cite{holden2016deep, martinez2017human, hernandez2019human, MOJO} largely ignores the presence of objects in the scene.

Modeling and synthesizing realistic whole-body grasping motions are challenging and remain unsolved due to a number of reasons. 
Firstly, whole-body grasping motions involve both full-body dynamics and dexterous finger movements~\cite{GRAB:2020}, while the high dimensional degrees of freedom make the synthesis of grasping motions complicated. 
Secondly, a whole-body grasping sequence exhibits complex and frequent body--scene and hand--object contacts which are challenging to synthesize in a perceptually realistic way. %
For example, the hand's surface should conform naturally to the object and there should be no foot-skating artifacts in the whole-body motion.
Thirdly, given only a starting pose and a target object in 3D space, there could be an infinite number of ways for the person to approach and grasp the object. The diversity of plausible grasping motions is further amplified by the large potential variation in object shape and pose. How to build an effective generative model that can capture this diversity and synthesize diverse realistic motions to grasp various 3D objects is an unsolved question. %

To address these challenges, we propose \textbf{SAGA} (\textbf{S}toch\textbf{A}stic whole-body \textbf{G}rasping with cont\textbf{A}ct), a novel whole-body grasping generation framework that can synthesize stochastic motions of a 3D human body approaching and grasping 3D objects. 
Our solution consists of two components: (1) a novel 3D body generator that synthesizes diverse static whole-body grasping end poses, and (2) a novel human motion generator that creates diverse and plausible motions between given start and end poses. 
We present two key insights on both components. First, instead of directly using parametric body models (e.g.~SMPL~\cite{SMPL-X:2019}) to represent 3D bodies, we employ the markers-based representation~\cite{MOJO} which captures 3D human shape and pose information with a set of sparse markers on the human body surface. 
As demonstrated in~\cite{MOJO}, the markers-based representation is easier for neural networks to learn than the latent parameters of the parametric body models, yielding more realistic motion. We show that the markers-based representation is especially advantageous to the latent body parameters for learning whole-body grasping, as the accumulation of errors along the kinematic chain has a significant impact on the physical plausibility of generated hand grasps, resulting in severe hand--object interpenetration.
Second, {\it contact} plays a central role in our pipeline. As the human moves in 3D space and grasps a 3D object, physical contact is key for modeling realistic motions and interactions. For both components of our method, we learn contact representations from data and use them to guide interaction synthesis, greatly improving the realism of the generated motion.%

For the static grasping pose generation,  we built a multi-task conditional variational autoencoder (CVAE) to jointly learn whole-body marker positions and fine-grained marker--object contact labels. During inference, given a target object in 3D space, our model jointly generates a diverse set of consistent full-body marker locations and the contact labels on both the body markers and the object surface. A contact-aware pose optimization module further recovers a parametric body mesh from the predicted markers, while explicitly enforcing the
hand--object contact by leveraging the predicted contact labels. 
Next, given the generated static whole-body grasping pose as the end pose, and an initial pose as a start pose, we propose a novel generative motion infilling network to capture motion
uncertainty and generate diverse motions in between. We design a CVAE-based architecture to generate both the diverse in-between motion trajectories and the diverse in-between local pose articulations. In addition, contacts between feet and the ground are also predicted as a multi-task learning objective to enforce a better foot-ground interaction. Furthermore, leveraging the predicted human--object contacts, we design a contact-aware motion optimization module to produce realistic grasp-oriented whole-body motions from the generated marker sequences.
By leveraging the GRAB~\cite{GRAB:2020} and AMASS~\cite{AMASS:ICCV:2019} datasets to learn our generative models, 
our method can successfully generate realistic and diverse whole-body grasping motion sequences for approaching and grasping a variety of 3D objects.

\myparagraph{Contributions.} In summary, we provide (1) a novel generative framework to synthesize diverse and realistic whole-body motions approaching and grasping various unseen objects for 3D humans that exhibit various body shapes,
(2) a novel multi-task learning model to jointly learn the static whole-body grasping poses and the body--object interactions, 
(3) a novel generative motion infilling model that can stochastically infill both the global trajectories and the local pose articulations, yielding diverse and realistic full-body motions between a start pose and end pose. We perform extensive experiments to validate technical contributions. Experimental results demonstrate both the efficacy of our full pipeline and the superiority of each component to existing solutions.

\section{Related Work}
\label{sec:related}
\myparagraph{Human Grasp Synthesis} is a challenging task and has been studied in computer graphics~\cite{rijpkema1991computer,pollard2005physically,10.1145/1141911.1141969,4293017,kalisiak2001grasp,Zhang2021ManipNetNM} and robotics~\cite{4293017, 6225086, 5654380, hsiao2006imitation, 5509126, liu2019generating}. With the advancement in deep learning, recent works also approach the realistic 3D human grasp synthesis task by leveraging large-scale datasets~\cite{GrapingField:3DV:2020, GRAB:2020, jiang2021graspTTA, brahmbhatt2019contactgrasp, Zhang2021ManipNetNM}, however they only focus on hand grasp synthesis.

Grasping Field~\cite{GrapingField:3DV:2020} proposes an implicit representation of hand--object interaction and  builds a model to generate plausible human grasps. GrabNet~\cite{GRAB:2020} proposes a CVAE to directly sample the MANO~\cite{MANO:SIGGRAPHASIA:2017} hand parameters, and additionally train a neural network to refine the hand pose for a more plausible hand--object contact. GraspTTA~\cite{jiang2021graspTTA} suggests using consistent hand--object interactions to synthesize realistic hand grasping poses. It sequentially generates coarse hand grasping poses and estimates consistent object contact maps, and using the estimated object contact maps, the produced hand pose is further optimized for realistic hand--object interactions. Similarly, ContactOpt~\cite{grady2021contactopt} proposes an object contact map estimation network and a contact-based hand pose optimization module to produce realistic hand--object interaction. Different from GraspTTA and ContactOpt, which predict consistent hand pose and hand--object contacts sequentially in two stages, we build a multi-task generative model that generates consistent whole-body pose and the mutual human-object contacts jointly to address a more complicated whole-body grasping pose learning problem. Going beyond the static grasp pose generation, provided the wrist trajectory and object trajectory, ManipNet~\cite{Zhang2021ManipNetNM} generates dexterous finger motions to manipulate objects using an autoregressive model. Nonetheless, to our best knowledge, none of the previous works studied 3D human whole-body grasp learning and synthesis.

\noindent
\textbf{3D Human Motion Synthesis.}
In recent years, human motion prediction has received a lot of attention in computer vision and computer graphics~\cite{fragkiadaki2015recurrent, jain2016structural, mao2019learning, wang2019imitation, chiu2019action, starke2020local, holden2017phase, ling2020character,cai2020learning, MOJO, liu2018learning, zhang2020learning}. 
Existing motion prediction models can also be split into two categories: deterministic~\cite{yan2019convolutional, hernandez2019human, kaufmann2020convolutional, holden2016deep, martinez2017human} and stochastic~\cite{yan2018mt, barsoum2018hp, li2021task, cai2021unified}. %
For deterministic motion prediction, \cite{kaufmann2020convolutional} adopt convolutional models to provide spatial or temporal consistent motions, and \cite{martinez2017human} propose an RNN with residual connections and sampling-based loss to model human motion represented by joints. For stochastic motion prediction, %
recently, Li \textit{et al}. \cite{li2021task} and Cai \textit{et al}. \cite{cai2021unified} use VAE-based models to address general motion synthesis problems. 
While these methods make great contributions to human motion understanding, they do not study the interaction with the 3D environment.

There are several works predict human motion paths or poses in scene context \cite{alahi2014socially, gupta20113d, savva2016pigraphs, tan2018and, gupta2018social, li2019putting, sadeghian2019sophie, makansi2019overcoming, tai2018socially, starke2019neural, helbing1995social, wang2021synthesizing, rempe2021humor, lemo2021, cao2020long, harvey2020robust}. 
Cao \textit{et al}. \cite{cao2020long} estimate goals, 3D human paths, and pose sequences given 2D pose histories and an image of the scene. However, the human is represented in skeletons, thus it is hard to accurately model body--scene contacts, which limits its application. Recently, Wang \textit{et al}. \cite{wang2021synthesizing} propose a pipeline to infill human motions in 3D scenes, which first synthesizes sub-goal bodies, then fills in the motion between these sub-goals, then refines the bodies. However, the generated motion appears unnatural especially in the foot--ground contact. \cite{harvey2020robust} presents an RNN-based network with contact loss and adversarial losses to handle motion in-betweening problems. They use the humanoid skeleton as the body representation and require 10 start frames and one end frame as input. %
\cite{rempe2021humor} adopts a conditional variational autoencoder to correct the pose at each timestamp to address noise and occlusions. They also use a test-time optimization to get more plausible motions and human--ground contacts. \cite{lemo2021} propose a contact-aware motion infiller %
to generate the motion of unobserved body parts. They predict motion with better foot--ground contact, but their deterministic model does not capture the nature of human motion diversity. Unlike the methods mentioned above, our generative motion infilling model, when given the first and the last frame, captures both the global trajectory diversity in between and the diversity of local body articulations.

\noindent
\textbf{Concurrent work.} GOAL~\cite{GOAL} builds a similar two-stage pipeline to approach the whole-body grasping motion generation, producing end pose first and then infilling the in-between motion. Unlike our work which captures both the diversity of grasping ending poses and in-between motions, however, GOAL builds a deterministic auto-regressive model to in-paint the in-between motion, which does not fully explore the uncertainty of grasping motions.

\section{Method}
\label{sec:method}

\myparagraph{Preliminaries.}
\textbf{(a) 3D human body representation.} (1) SMPL-X~\cite{SMPL-X:2019} is a parametric human body model which models body mesh with hand details. In this work, the SMPL-X body parameters $\bm{\Theta}$ include the shape parameters $\bm{\beta} \in \mathbb{R}^{10}$, the body global translation $\bm{t} \in \mathbb{R}^3$, the 6D continuous representation~\cite{Zhou_2019_CVPR} of the body rotation $\bm{R} \in \mathbb{R}^6$, and full-body pose parameters $\bm{\theta} = [\bm{\theta_b}, \bm{\theta_h}, \bm{\theta_e}]$, where $\bm{\theta_b} \in \mathbb{R}^{32}$, $\bm{\theta_h} \in \mathbb{R}^{48}$, $\bm{\theta_e} \in \mathbb{R}^{6}$ are the body pose in the Vposer latent space~\cite{SMPL-X:2019}, the hands pose in the MANO~\cite{MANO:SIGGRAPHASIA:2017} PCA space and the eyes pose, respectively; %
(2) Markers-based representation~\cite{MOJO} captures the body shape and pose information with the 3D locations $\bm{M} \in \mathbb{R}^{N\times3}$ of a set of sparse markers on the body surface, where $N$ is the number of markers. %
We learn the markers representation in our neural networks, from which we further recover SMPL-X body mesh.  
\textbf{(b) 3D objects} are represented with centered point cloud data $\bm{O}$ and the objects height $t_{\bm{O}} \in \mathbb{R}^1$. We sample 2048 points on the object surface and each point has 6 features (3 XYZ positions + 3 normal features).

\myparagraph{Notations.} For clarity, in the following text, $\Tilde{X}$ and $\hat{X}$ denote the CVAE reconstruction result of $X$, and random samples of $X$ from CVAE, respectively.

\subsection{Overview}
\begin{figure*}[t]
    \centering
    \includegraphics[width=1\linewidth]{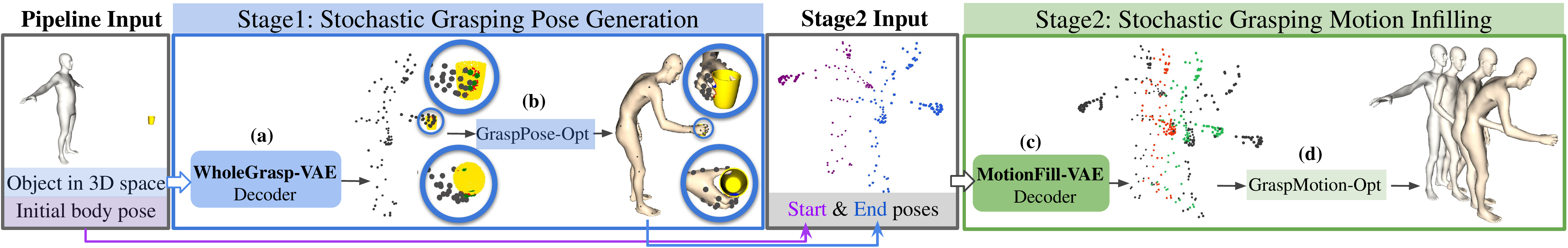}
    \caption{Illustration of our two-stage pipeline. Given an object in 3D space and a start pose, our method produces diverse human whole-body grasping motions. In stage 1, (a) given 3D object information, our WholeGrasp-VAE (\S~\ref{WholeGrasp}) decoder generates whole-body grasping poses represented by markers and mutual marker--object contact probabilities (green markers and red areas on the object surface indicate high contact probabilities); (b) GraspPose-Opt (\S~\ref{WholeGrasp}) further recovers body mesh from predicted markers. We use the generated grasping pose as the targeted end pose. Then in stage 2, (c) we feed in the start pose and the end pose into the MotionFill-VAE decoder (\S~\ref{MotionInfilling}) to generate the in-between motions in marker representation, and (d) and GraspMotion-Opt (\S~\ref{fullpipeline}) further recovers smooth and realistic whole-body grasping motions. %
    }
    \label{fig:pipeline}
    \vspace{-1em}
\end{figure*}

Given an initial human pose and a 3D object randomly placed in front of the human within a reasonable range, our goal is to generate realistic and diverse whole-body motions, starting from the given initial pose and approaching to grasp the object. As presented in Fig.~\ref{fig:pipeline}, we propose a two-stage grasping motion generation pipeline to approach this task.

\myparagraph{Stage 1: Stochastic whole-body grasping ending pose generation (\S~\ref{WholeGrasp}).} We first build an object-conditioned multi-task CVAE which synthesizes whole-body grasping ending poses in markers and the explicit human--object contacts. We further perform contact-aware pose optimization to produce 3D body meshes with realistic interactions with objects by leveraging the contacts information. %

\myparagraph{Stage 2: Stochastic grasp-oriented motion infilling.} We build a novel generative motion infilling model (\S~\ref{MotionInfilling}) which takes the provided initial pose and the generated end pose in stage 1 as inputs, and outputs diverse intermediate motions. We further process the generated motions via a contact-aware optimization step (\S~\ref{fullpipeline}) to produce realistic human whole-body grasping motions.

\vspace{-0.8em}
\subsection{Whole-Body Grasping Pose Generation}
\label{WholeGrasp}
To synthesize diverse whole-body poses to grasp a given object, we propose a novel multi-task WholeGrasp-VAE to learn diverse yet consistent grasping poses and mutual contacts between human and object. The explicit human--object contacts provide fine-grained human--object interaction information which helps to produce realistic body meshes with high-fidelity interactions with the object. %

\begin{figure}[t]
  \centering
  \includegraphics[width=0.7\linewidth]{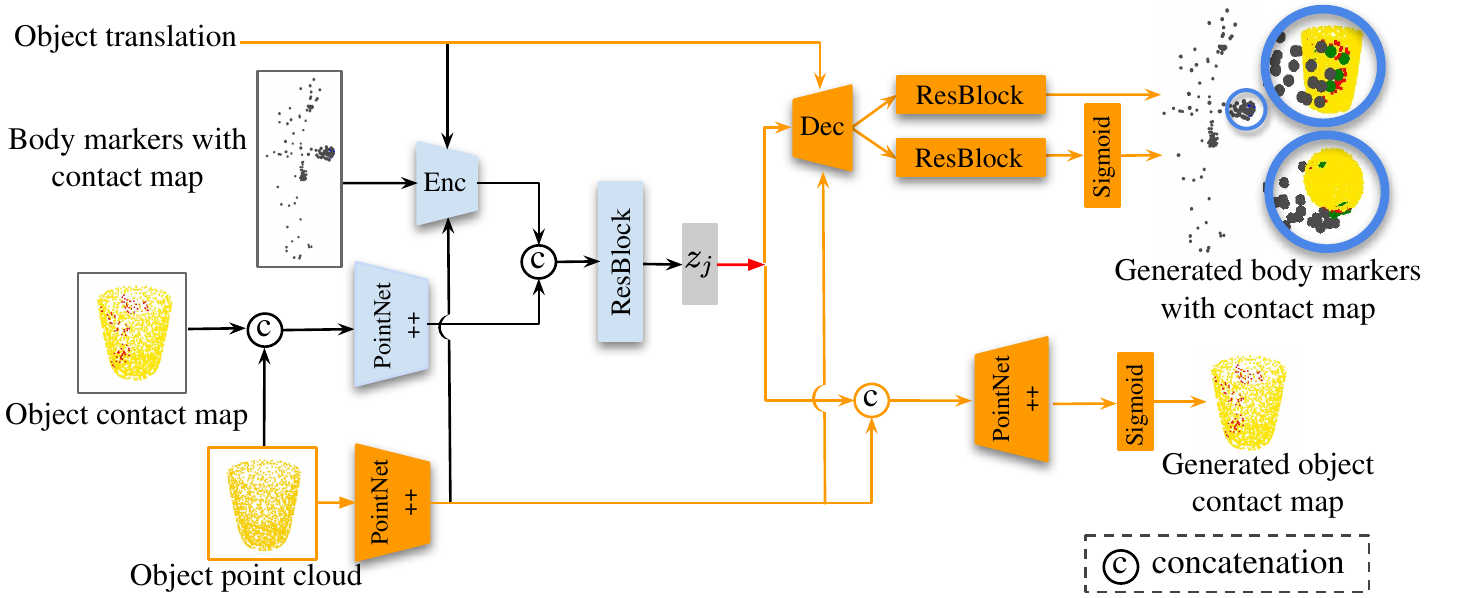}
   \caption{The WholeGrasp-VAE design. WholeGrasp-VAE jointly learns the (1) body marker locations; (2) body marker contacts (markers with high contact probability are shown in green); (3) object contact map (the area with high contact probability is shown in red). The red arrow indicates sampling from the latent space. At inference time, activated modules are shown in {\color{orange} orange}.}
   \label{fig:Arch-WholeGraspCVAE}
   \vspace{-1em}
\end{figure}

\myparagraph{Model Architecture.} We visualize the multi-task WholeGrasp-VAE design in Fig.~\ref{fig:Arch-WholeGraspCVAE}. The encoder takes the body markers' positions $\bm{M} \in \mathbb{R}^{N\times3}$, body markers contacts $C_{\bm{M}} \in \{0, 1\}^{N}$ and object contacts $C_{\bm{O}} \in \{0, 1\}^{2048}$ as inputs, where $N$ is the number of markers, and learns a joint Gaussian latent space $\mathbf{z_j}$. We use PointNet++ \cite{qi2017pointnet++} to encode the object feature.%

\myparagraph{Training.} The overall training objective is given by $\mathcal{L}_{train} = \mathcal{L}_{rec} + \lambda_{KL}\mathcal{L}_{KL} + \lambda_c \mathcal{L}_{c}$, where $\lambda_{KL}, \lambda_c$ are hyper-parameters.

\textbf{Reconstruction loss} includes the L1 reconstruction loss of body markers' postions and the binary cross-entropy (BCE) loss of contact probabilities:
\begin{small}
\begin{align}
    \mathcal{L}_{rec} = |\bm{M} - \bm{\Tilde{M}}| + \lambda_{\bm{M}}\mathcal{L}_{bce}(C_{\bm{M}}, \Tilde{C}_{\bm{M}}) 
    + \lambda_{\bm{O}}\mathcal{L}_{bce}(C_{\bm{O}}, \Tilde{C}_{\bm{O}}).
    \label{eq:train-1-rec-loss}
\end{align}
\end{small}
\vspace{-0.5em}

\textbf{KL-divergence loss.} We employ the robust KL-divergence term~\cite{MOJO} to avoid the VAE posterior collapse:
\begin{small}
\begin{align}
    \mathcal{L}_{KL} = \Psi(D_{KL}(q(\mathbf{z_j}|\bm{M}, C_{\bm{M}}, C_{\bm{O}}, t_{\bm{O}}, \bm{O})|| \mathcal{N}(\mathbf{0}, \mathbf{I}))),
    \label{eq:train-s1-kl}
\end{align}
\end{small}

\vspace{-0.5em}
\noindent
where $\Psi(s) = \sqrt{s^2+1}-1$ ~\cite{charbonnier}.
This function automatically penalizes the gradient to update the above KLD term, when the KL-divergence is small.

\textbf{Consistency loss.} We use a consistency loss to implicitly encourage consistent predictions of marker positions and mutual marker--object contacts:
\begin{small}
\begin{align}
    \mathcal{L}_{c} = \sum_{m \in \bm{M}, \Tilde{m} \in \bm{\Tilde{M}}} \Tilde{C}_m  |d(\Tilde{m}, \bm{O}) - d(m, \bm{O})| %
    + \sum_{o \in \bm{O}} \Tilde{C}_o  |d(o, \bm{\Tilde{M}}) - d(o, \bm{M})|,
\end{align}
\end{small}

\vspace{-0.5em}
\noindent
\scalebox{0.9}{$d(x, \mathcal{Y})=\min\limits_{y\in\mathcal{Y}}||x-y||^2_2$} is the minimum distance from point $x$ to point cloud \scalebox{0.8}{$\mathcal{Y}$}.

\vspace{-0.6em}
\myparagraph{Inference.} %
During inference, we feed the provided target object information into the WholeGrasp-VAE decoder to generate plausible body markers $\hat{\bm{M}}$ and marker--object contact labels $\hat{C}_{\bm{M}}$, $\hat{C}_{\bm{O}}$. We design a contact-aware pose optimization algorithm, GraspPose-Opt, 
to generate a realistic body mesh from markers and refine body pose for high-fidelity human--object interaction by leveraging the fine-grained human--object contacts. Specifically, by optimizing SMPL-X parameters $\bm{\Theta}$, the overall optimization objective is given by:
\begin{small}
\begin{align}
    E_{opt}(\bm{\Theta}) = E_{fit}  + E_{colli}^o + E_{cont}^o + E_{cont}^g.
    \label{eq:opt-1}
\end{align}
\end{small}

\vspace{-0.5em}
\textbf{Marker fitting loss.} To project the predicted markers to a valid body mesh, we minimize the L1 distance between the sampled markers $\hat{\bm{M}}$ and the queried markers $\bm{M}(\bm{\Theta})$ on the SMPL-X body mesh:
\begin{small}
\begin{align}
    E_{fit}(\bm{\Theta}) = |\hat{\bm{M}} - \bm{M}(\bm{\Theta})| + \alpha_{\bm{\theta}}|\bm{\theta}|^2,%
    \label{eq:opt-1-fit}
\end{align}
\end{small}

\vspace{-0.5em}
\noindent
where $\alpha_{\bm{\theta}}$ is the pose parameters regularization weight.

\textbf{Object contact loss.} By leveraging sampled contact maps, we propose a mutual contact loss to encourage body markers and object points with high contact probabilities to contact the object surface and body surface, respectively.
\begin{small}
\begin{align}
    E_{cont}^o(\bm{\Theta}) = \alpha_{cont}^{o}\sum_{o\in\bm{O}}\hat{C}_od(o, \mathcal{V}_{\bm{B}}(\bm{\Theta})) %
    + \alpha_{cont}^{m}\sum_{m\in \bm{M}(\bm{\Theta})}\hat{C}_md(m, \bm{O}). 
    \label{eq:opt-1-contact} 
\end{align}
\end{small}

\vspace{-0.5em}
\noindent
where $\mathcal{V}_{\bm{B}}(\bm{\Theta})$ denotes the SMPL-X body vertices.

\textbf{Collision loss.} We employ a signed-distance based collision loss to penalize the body--object interpenetration:
\begin{small}
\begin{align}
    E_{colli}(\bm{\Theta}) = \alpha_{colli}^{\bm{B}}\sum_{b\in{\mathcal{V}_{\bm{B}}^h(\bm{\Theta})}}\max(-\mathcal{S}(b, \bm{O}), \sigma_b) 
    + \alpha_{colli}^{\bm{O}}\sum_{o\in \bm{O}}\max(-\mathcal{S}(o, \mathcal{V}_{\bm{B}}^h(\bm{\Theta})), \sigma_o)
    \label{eq:opt-1-colli}
\end{align}
\end{small}

\vspace{-0.5em}
\noindent
where $\mathcal{S}(x, \mathcal{Y})$ is the signed-distance from point $x$ to point cloud $\mathcal{Y}$, $\mathcal{V}_{\bm{B}}^h(\bm{\Theta})$ denotes the hand vertices, and $\sigma_b$, $\sigma_o$ are small interpenetration thresholds.

\textbf{Ground contact loss} is given by \scalebox{0.9}{$E_{cont}^g(\bm{\Theta}) = \alpha_{cont}\sum_{v \in \mathcal{V}^f_{\bm{{B}}}}|h(v)|$}, where we penalize the heights of feet vertices \scalebox{0.9}{$\mathcal{V}^f_{\bm{{B}}}$} to enforce a plausible foot--ground contact.

\begin{figure}[t]
  \centering
  \includegraphics[width=0.75\linewidth]{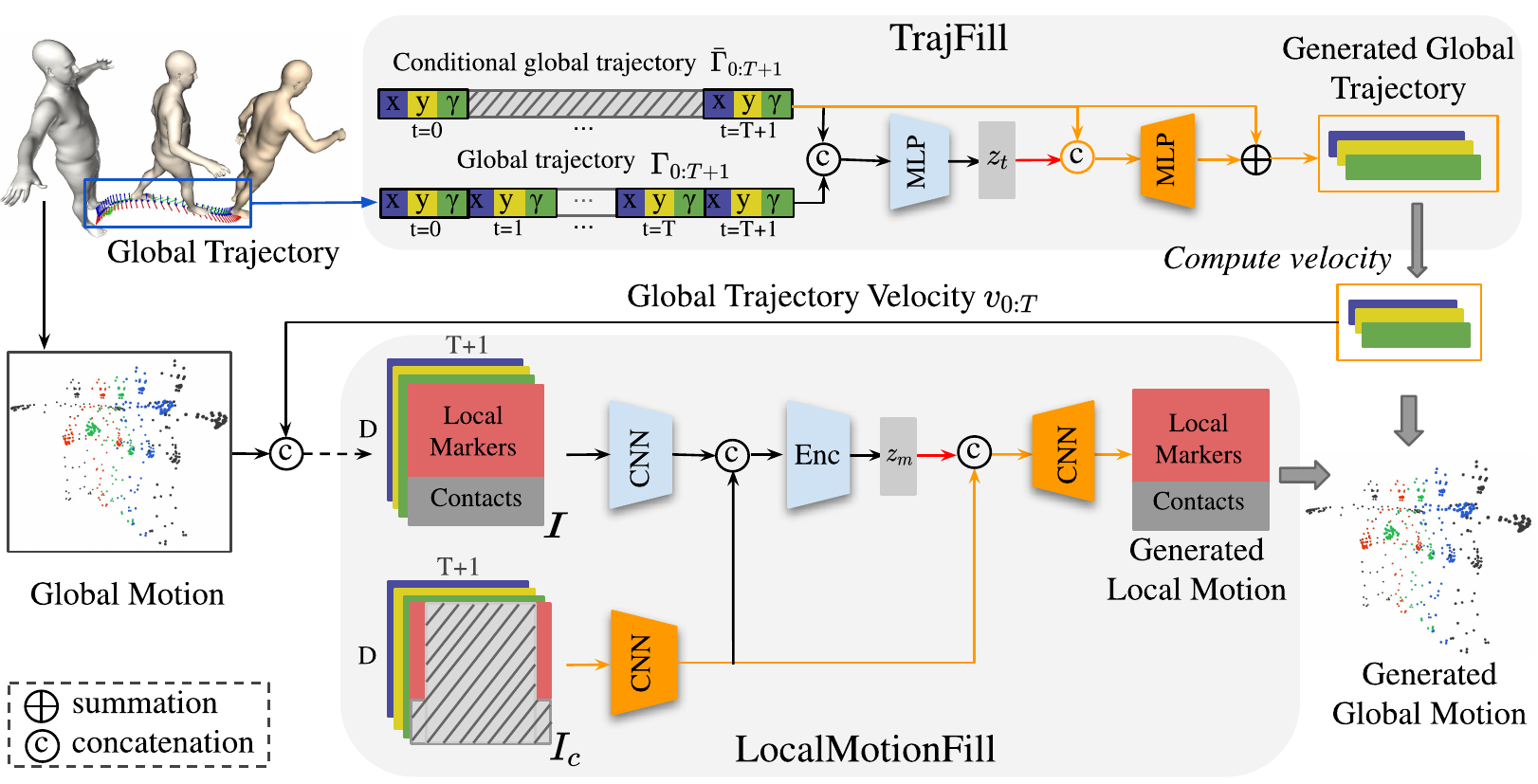}
   \caption{MotionFill-VAE consists of two concatenated CVAEs: (1) TrajFill outputs the infilled global root trajectory when the start root and the end root are given; (2) LocalMotionFill takes the global trajectory information from TrajFill as one of the inputs, and it outputs the infilled local motion when the start pose, the end pose, and the global trajectory are given. We reconstruct the global motion from the generated global trajectory and the local motion. The red arrow indicates sampling from the latent space. The dash arrow indicates the input processing step for building the four-channel motion image (one local motion channel with contact states and three root velocity channels). At inference time, activated modules are shown in {\color{orange} orange}.}
   \label{fig:Arch-MotionFillCVAE}
  \vspace{-1em}
\end{figure}
\subsection{Generative Motion Infilling}
\label{MotionInfilling}

Given body markers on the start and end poses produced by GraspPose-Opt, {\emph{i.e.}}, $\bm{M}_{0}$ and $\bm{M}_{T}$, many in-between motions are plausible. To model such uncertainty, we build a novel generative motion infilling model, namely MotionFill-VAE, to capture both the uncertainties of intermediate global root (pelvis joint) trajectories and intermediate root-related local body poses. Specifically, given motion $\bm{M}_{0:T}$ represented in a sequence of markers positions, following~\cite{holden2016deep, kaufmann2020convolutional, lemo2021}, we represent the global motion $\bm{M}_{0:T}$ with a hierarchical combination of global root velocity $\bm{v}_{0:T}$ (where $\bm{v}_{t}=\bm{\Gamma}_{t+1}-\bm{\Gamma}_{t}, t \in [0, T]$, $\bm{\Gamma}$ and $\bm{v}$ denote the root trajectory and root velocity respectively) and the trajectory-conditioned local motion $\bm{M}^l_{0:T}$. Accordingly, we build the MotionFill-VAE to capture both the conditional global trajectory distribution $P(\bm{\Gamma}_{0:T+1}|\bm{\Gamma}_0, \bm{\Gamma}_T)$ and the conditional local motion distribution $P(\bm{M}^l_{0:T}|\bm{v}_{0:T}, \bm{M}_0^l, \bm{M}_T^l)$.

\myparagraph{Model Architecture.} As shown in Fig.~\ref{fig:Arch-MotionFillCVAE}, the MotionFill-VAE consists of two concatenated CVAEs: \textbf{(1) TrajFill} learns the conditional intermediate global root trajectory latent space $\bm{z_t}$. Taking the root states $\bm{\Gamma}_0$ and $\bm{\Gamma}_T$ as inputs, which are derived from the given start and end pose, our goal is to get the trajectory $\bm{\Gamma}_{0:T+1}$. Instead of directly learning $\bm{\Gamma}_{0:T+1}$, we build TrajFill to learn the trajectory deviation $\Delta \bm{\Gamma}_{0:T+1} = \bm{\Gamma}_{0:T+1} - \overline{\bm{\Gamma}}_{0:T+1}$, where $\overline{\bm{\Gamma}}_{0:T+1}$ is a straight trajectory which is a linear interpolation and one-step extrapolation of the given $\bm{\Gamma}_0$ and $\bm{\Gamma}_T$. We further compute the velocity $\bm{v}_{0:T}$ from the predicted trajectory $\bm{\Gamma}_{0:T+1}$. \textbf{(2) LocalMotionFill} learns the conditional intermediate local motion latent space $\bm{z_m}$. Taking the TrajFill output $\bm{v}_{0:T}$ and the given $\bm{M}_0, \bm{M}_T$ as inputs, LocalMotionFill generates the trajectory-conditioned local motion sequence. Specifically, following~\cite{lemo2021}, we build a four-channel image $\bm{I}$, which is a concatenation of local motion information with foot--ground contact labels and root velocity, and we use it as the input to our CNN-based LocalMotionFill architecture. Similarly, we build the four-channel conditional image $\bm{I}_c$ with the unknown motion in between filled with all 0.

\myparagraph{Training.} 
The training loss is \scalebox{0.9}{$\mathcal{L}_M = \mathcal{L}_{rec}+\lambda_{KL} \mathcal{L}_{KL}$}, and \scalebox{0.9}{$\lambda_{KL}$} is hyper-parameter.

\textbf{Reconstruction loss} $\mathcal{L}_{rec}$ contains the global trajectory reconstruction, local motion reconstruction and foot--ground contact label reconstruction losses: 
\begin{small}
\begin{align}
\mathcal{L}_{rec} &= \sum_{t=0}^{T+1}|\bm{\Gamma}_{t}-\bm{\Tilde{\Gamma}}_{t}| + \lambda_1\sum_{t=0}^{T}|\bm{v}^{\bm{\Gamma}}_{t}-\bm{\Tilde{v}}_{t}^{\bm{\Gamma}}| + \lambda_2\mathcal{L}_{bce}(C_F, \Tilde{C}_F) \nonumber \\
&+ \lambda_3\sum_{t=0}^{T}|\bm{M}_{t}^l-\bm{\Tilde{M}}_{t}^{l}| 
+ \lambda_4\sum_{t=0}^{T-1}|\bm{v}^{\bm{M^l}}_{t}-\bm{\Tilde{v}^{M^l}}_{t}|,
\label{eq:motionfill-rec}
\end{align}
\end{small}
where \scalebox{0.9}{$\bm{v_t^{(*)}} = \bm{(*)}_{t+1}-\bm{(*)}_{t}$} denotes the velocity, and $\lambda_1 - \lambda_4$ are hyper-parameters.

\textbf{KL-divergence loss.} We use the robust KL-divergence loss for both TrajFill and LocalMotionFill: 
\begin{small}
\begin{align}
    \mathcal{L}_{KL} &= \Psi(D_{KL}(q(\bm{z}_{t}|\bm{\Gamma}_{0:T+1}, \overline{\bm{\Gamma}}_{0:T+1})||\mathcal{N}(\mathbf{0}, \mathbf{I}))) 
    +\Psi(D_{KL}(q(\bm{z}_{m}|\bm{I}, \bm{I}_c)|| \mathcal{N}(\mathbf{0}, \mathbf{I}))).
\label{eq:motionfill-kl}
\end{align}
\end{small}

\myparagraph{Inference.} At inference time, given the start and end body markers \scalebox{0.9}{$\bm{M}_0, \bm{M}_T$} with known root states \scalebox{0.9}{$\bm{\Gamma}_0, \bm{\Gamma}_T$}, by first feeding the initial interpolated trajectory \scalebox{0.9}{$\overline{\bm{\Gamma}}_{0:T+1}$} into the decoder of TrajFill, we generate stochastic in-between global motion trajectory \scalebox{0.9}{$\hat{\bm{\Gamma}}_{0:T+1}$}. Next, with the given \scalebox{0.9}{$\bm{M}_0, \bm{M}_T$} and the generated \scalebox{0.9}{$\hat{\bm{\Gamma}}_{0:T+1}$}, we further build the condition input image $\bm{I}_c$ as the input to the LocalMotionFill decoder, from which we can generate infilled local motion sequences \scalebox{0.9}{$\hat{\bm{M}}_{0:T}^l$} and also the foot--ground contact probabilities \scalebox{0.9}{$\hat{C}_{F_{0:T}}$}. Finally, we reconstruct the global motion sequences \scalebox{0.9}{$\hat{\bm{M}}_{0:T}$} from the generated \scalebox{0.9}{$\hat{\bm{\Gamma}}_{0:T}$} and \scalebox{0.9}{$\hat{\bm{M}}_{0:T}^l$}.

\subsection{Contact-aware Grasping Motion Optimization}
\label{fullpipeline}
With the generated marker sequences $\hat{\bm{M}}_{0:T}$, foot--ground contacts $\hat{C}_F$ from MotionFill-VAE, and the human--object contacts $\hat{C}_{\bm{M}}, \hat{C}_{\bm{O}}$ from WholeGrasp-VAE, we design GraspMotion-Opt, a contact-aware motion optimization algorithm, to recover smooth motions $\bm{B}_{0:T}$ with natural interactions with the scene.

Similar to GraspPose-opt, we propose the contact-aware marker fitting loss:
\begin{equation}
    \begin{small}
        E_{basic}(\bm{\Theta}_{0:T}) = \sum_{t=0}^{T}(E_{fit}(\bm{\Theta}_t) + E_{colli}^o(\bm{\Theta}_t)) 
        +  \sum_{t=T-4}^{T}E_{cont}^o(\bm{\Theta}_t),
        \label{eq:opt-2-basic}
    \end{small}
\end{equation}

\noindent
where $E_{fit}, E_{cont}^o, E_{colli}^o$ are formulated in Eq.~\ref{eq:opt-1-fit}-\ref{eq:opt-1-colli}, and we only apply object contact loss $E_{colli}^o$ on the last 5 frames.

We design the following loss to encourage a natural hand grasping motion by encouraging the palm to face the object's surface on approach.
\begin{small}
\begin{align}
    E_{g}(\bm{\Theta}_{0:T}) =& \sum_{t=0}^T\alpha_t\sum_{m \in \mathcal{V}^p_{\bm{B}}(\bm{\Theta}_t)}\mathbbm{1}(d(m, \bm{O})<\sigma)(\cos\gamma_m-1)^2,
    \label{eq:opt-2-grasp}
\end{align}
\end{small}

\noindent
where $\alpha_t=1-(\frac{t}{T})^2$, $\mathcal{V}^p_{\bm{B}}(\bm{\Theta})$ denotes the selected vertices on palm, and $\gamma_m$ is the angle between the palm normal vector and the vector from palm vertices to the closest object surface points. We only apply this constraint when palm vertices are close to the object's surface (within radius $\sigma=1$cm).

Inspired by~\cite{lemo2021}, we enforce smoothness on the motion latent space to yield smoother motion, and we also reduce the foot skating artifacts by leveraging the foot--ground contact labels $\hat{C}_F$. For more details, please refer to the Appendix.

\section{Experiments}
\textbf{Datasets.} (1) We use \textbf{GRAB}~\cite{GRAB:2020} dataset to train and evaluate our WholeGrasp-VAE and also finetune the MotionFill-VAE. For WholeGrasp-VAE training and evaluation, following \cite{GRAB:2020}, we take all frames with right-hand grasps and have the same train/valid/test set split. For MotionFill-VAE training, we downsample the motion sequences to 30fps and clip 62 frames per sequence, with last frames being in stable grasping poses. (2) We use the \textbf{AMASS}~\cite{AMASS:ICCV:2019} dataset to pretrain our LocalMotionFill-CVAE. We down-sample the sequences to 30 fps and cut them into clips with 61 frames. (3) We take unseen objects from \textbf{HO3D}~\cite{hampali2020honnotate} dataset to test the generalization ability of our method.%

We conduct extensive experiments to study the effectiveness of each stage in our pipeline. In Sec.~\ref{exp-wholegrasp} and Sec.~\ref{exp-motionfill}, we study our static grasping pose generator and the stochastic motion infilling model respectively. In Sec.~\ref{exp-fillpipeline}, we evaluate the entire pipeline performance for synthesizing stochastic grasping motions. We encourage readers to watch the \href{https://jiahaoplus.github.io/SAGA/saga.html}{video} of generated grasping poses and motions.
\label{sec:experiment}

\subsection{Stochastic Whole-body Grasp Pose Synthesis}
\label{exp-wholegrasp}
We evaluate our proposed stochastic whole-body grasp pose generation module on GRAB dataset. We also conduct ablation studies to study the effectiveness of several proposed components, including the multi-task CVAE design and the contact-aware optimization module design.
\begin{table}[t]
    \centering
    \caption{Comparisons with the extended GrabNet baseline and ablation study result on the multi-task WholeGrasp-VAE design. Numbers in {\textbf{bold}}/{\color{blue}{blue}} indicates the {\textbf{best}}/{\color{blue}{second-best}} respectively.}
    \resizebox{\columnwidth}{!}{
    \begin{threeparttable}
    \begin{tabular}{lcccc}
    \toprule
       \multirow{2}{*}{\textbf{Method}} & \textbf{APD} & \textbf{Contact Ratio} & \textbf{Inter. Vol.}  & \textbf{Inter. Depth}\\
         & $(\uparrow)$ & $(\uparrow)$ & $[cm^3]$ $(\downarrow)$ & $[cm]$ $(\downarrow)$ \\
    \midrule
    GrabNet~\cite{GRAB:2020}-SMPLX & 0.33 &0.65 & 14.15 & 0.78 \\
    \midrule
    WholeGrasp-single w/o opt.$^*$ & \multirow{2}{*}{{\textbf{2.94}}} & 0.90 & 11.44 & 0.78 \\
    WholeGrasp-single w/ heuristic opt. &  &  0.81 & {\textbf{0.21}} & {\textbf{0.12}}\\
    \midrule
    WholeGrasp w/o opt.$^*$ & \multirow{2}{*}{{\color{blue}{2.92}}}  & {\textbf{0.96}} &  12.20 & 0.85\\
    WholeGrasp w/ opt. (\textbf{Ours}) &  &  {\color{blue}{0.94}} & {\color{blue}{0.48}} & {\color{blue}{0.16}} \\
    \bottomrule
    \end{tabular}
    \begin{tablenotes}
 $^*$ Body meshes are recovered from sampled markers with only $E_{fit}$ in Eq.~\ref{eq:opt-1-fit}. 
\end{tablenotes}
     \end{threeparttable}
     }
    \label{tab:baseline-2}
    \vspace{-1em}
\end{table}

\myparagraph{Baseline.} GrabNet~\cite{GRAB:2020} builds a CVAE to generate the MANO hand parameters for grasping a given object, and we extend GrabNet to whole-body grasp synthesis by learning the whole-body SMPL-X parameters. We compare our method against the extended GrabNet (named as GrabNet-SMPLX)\footnote{Please refer to the Appendix for experimental setup and implementation details.\label{footnote_implementation}}.

\myparagraph{Evaluation Metrics.}\textit{(1) Contact Ratio.} To evaluate the grasp stability, we measure the ratio of body meshes being in minimal contact with object meshes. \textit{(2) Interpenetration Volume and Depth.} We measure the interpenetration volumes and depths between the body and object mesh. Low interpenetration volume and depth with a high contact ratio are desirable for perceptually realistic body--object interactions. \textit{(3) Diversity.} We follow~\cite{yuan2020dlow} to employ the Average L2 Pairwise Distance (\textbf{APD}) to evaluate the diversity within random samples.%

\myparagraph{Results.} In Table~\ref{tab:baseline-2}, we compare our method against the extended GrabNet baseline. Because the extended GrabNet baseline does not include an additional body mesh refinement step, we compare it to our results without GraspPose-Opt optimization (WholeGrasp w/o. opt. in Table~\ref{tab:baseline-2}). Our method w/o optimization outperforms the extended GrabNet baseline in the sample diversity (APD) and achieves higher contact ratio and smaller intersection. The extended GrabNet experiment demonstrates the challenges in learning the whole-body pose parameters for a plausible human--object interaction, with marker representation appearing to be more favorable for learning human grasping pose. Nevertheless, the derived body meshes from markers without pose optimization still have human--object interpenetration, and our contact-aware pose optimization (WholeGrasp w/ opt. in Table~\ref{tab:baseline-2}) drastically reduces the human--object collision issue while maintaining a high contact ratio. 
\begin{table}[t]
    \centering
    \caption{Ablation studies on different optimization losses ($E_{fit}, E_{colli}, E_{cont}^o$ in Eq.~\ref{eq:opt-1-fit}- Eq.~\ref{eq:opt-1-contact}). We fit ground truth markers (GT columns) and sampled markers (Samples columns), and numbers in {\textbf{bold}}/{\color{blue}{blue}} indicate the {\textbf{best}}/{\color{blue}{second-best}} respectively.}
    \begin{tabular}{lccccccc ccc}
    \toprule
          &  \multicolumn{2}{c}{Contact Ratio$(\uparrow)$} & & \multicolumn{2}{c}{Inter. Vol.$(\downarrow)$} & & \multicolumn{2}{c}{Inter. Depth$(\downarrow)$} \\
           \cline{2-3} \cline{5-6}  \cline{8-9}
          & GT & Samples && GT & Samples && GT & Samples \\
    \midrule
   \hspace{0.8cm} GT Mesh & 0.99&- && 2.04&- && 0.45 & - \\
    \midrule
    $E_{fit}+E_{cont}^g$ &  \textbf{0.99} & \textbf{0.96} && 2.21&12.20 && 0.46 & 0.85 \\
    $E_{fit}+E_{cont}^g + E_{colli}$ & 0.25 &0.24&& \textbf{0.12}&\textbf{0.12} && \textbf{0.04} & \textbf{0.07}\\
    $E_{fit}+E_{cont}^g + E_{colli} + E_{cont}^o$ & \color{blue}{0.94} &\color{blue}{0.94}&& \color{blue}{0.52}&\color{blue}{0.48} && \color{blue}{0.17} & \color{blue}{0.16}\\
    \bottomrule
    \end{tabular}
    \label{tab:ablation-losses}
\end{table}
\begin{figure}
    \centering
    \includegraphics[width=0.8\linewidth]{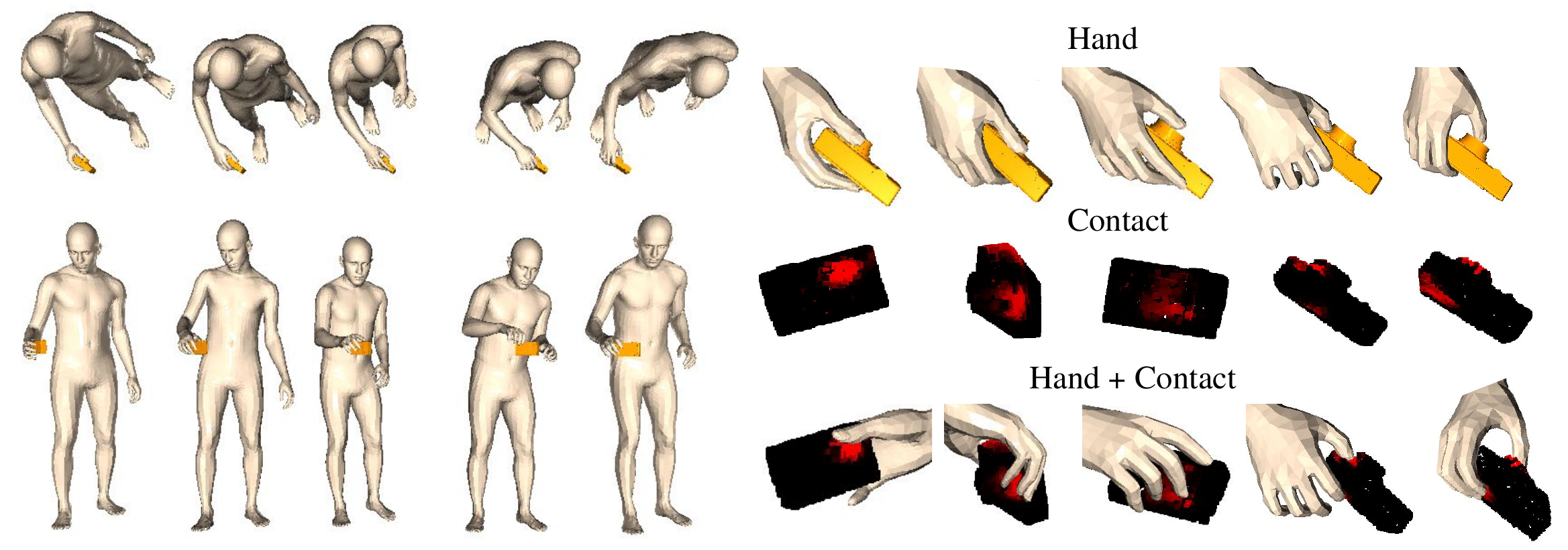}
    \caption{Five \textbf{random} samples for an unseen object placed at the same positions. Left side: top view and front view of generated whole-body poses. Right side: hand grasping details and generated object contact maps (red areas indicate high contact probability).}
    \label{fig:wholeGrasp-results-samples}
    \vspace{-1.2em}
\end{figure}

Fig.~\ref{fig:wholeGrasp-results-samples} presents 5 \textbf{random} samples together with the generated object contact maps and hand grasping details for an unseen object. We can see that our models generate natural grasping poses with diverse body shapes and whole-body poses.%

\myparagraph{Ablation Study.} \textit{(1) Multi-task WholeGrasp design:} To study the effect of learning human--object contact labels, we build a single-task WholeGrasp-VAE architecture which only learns the markers' positions (WholeGrasp-single in Table~\ref{tab:baseline-2}). A similar pose optimization step as our GraspPose-opt further refines the grasping pose (WholeGrasp-single w/ heuristic opt. in Table~\ref{tab:baseline-2}), but we replace the mutual contact loss $E_{cont}$ in Eq.~\ref{eq:opt-1-contact} with a heuristic contact loss which is based on a pre-defined hand contact pattern. Both the single-task and multi-task WholeGrasp experiments demonstrate the benefit of using contact to refine the human--object interaction, and our multi-task WholeGrasp with explicit mutual contact learning outperforms the single-task setup with the pre-defined hand contact pattern. \textit{(2) Study of GraspPose-Opt} (see Table~\ref{tab:ablation-losses}): We evaluate recovered body meshes from both the ground truth markers and the randomly sampled markers, and also the ground truth body mesh. By fitting the body mesh to ground truth markers, our proposed GraspPose-Opt with only $E_{fit}$ can recover an accurate body mesh with the human--object interaction metrics comparable to the ground truth mesh. The proposed $E_{colli}$ and $E_{cont}$ help to recover realistic human--object interaction significantly.%

\begin{table}[t]
    \centering
     \caption{Comparisons with motion infilling baselines. Best results are in boldface.}
    \setlength{\tabcolsep}{2mm}{
    \begin{threeparttable}
    \begin{tabular}{c l c c c c}
    \toprule
      & \multirow{2}{*}{Methods} & ADE  & Skat & \multicolumn{2}{c}{PSKL-J $(\downarrow)$} \\
    \cline{5-6}
  & & $(\downarrow)$ &  $(\downarrow)$ & (P, GT) & (GT, P) \\
    \midrule
   & CNN-AE \cite{kaufmann2020convolutional} & 0.091   & 0.245 & 0.804 & 0.739 \\
Local motion & LEMO \cite{lemo2021} & 0.083 & 0.152 & 0.507 & 0.447 \\
 infilling$^*$ &  PoseNet\cite{wang2021synthesizing} & 0.090 & 0.236 & 0.611 & 0.668 \\
  &  \textbf{Ours}-Local$^\dag$ & \textbf{0.079} & \textbf{0.137} & \textbf{0.377} & \textbf{0.327} \\
    \midrule
 Traj + local  & Route+PoseNet~\cite{wang2021synthesizing} & 0.219 & 0.575 & 0.955 & 0.884 \\
 motion infilling &  \textbf{Ours}$^\dag$ & \textbf{0.083} & \textbf{0.394} & \textbf{0.772} & \textbf{0.609} \\
    \bottomrule
    \end{tabular}
    \begin{tablenotes}
    \small
 $^*$ Ground truth trajectories are used in the local motion infilling experiments.\\
 $^\dag$ Generative models. And all the other methods are deterministic models.
\end{tablenotes}
     \end{threeparttable}
    }
    \label{tab:exp-motionfillvae}
    \vspace{-1em}
\end{table}

\subsection{Stochastic Motion Infilling}
\label{exp-motionfill}

We evaluate our motion infilling model on AMASS and GRAB datasets. To our best knowledge, we are the first generative model to learn both the global trajectory and the local motion infilling given only one start pose and end pose. We compare our method with several representative motion infilling models.

\myparagraph{Baselines.} Wang \textit{et al.}~\cite{wang2021synthesizing} proposed two sequential yet separate LSTM-based deterministic networks to first predict global trajectory (RouteNet) and then the local pose articulations (PoseNet) to approach the motion infilling task, and we take this sequential network (named as Route+PoseNet) as a baseline to our end-to-end generative global motion infilling model. There are some existing works which take the ground truth trajectory, start pose and end poses as inputs to predict the intermediate local poses, and following the same task setup, we also compare the generative local motion infilling component in our network against these baselines, including the convolution autoencoder network (CNN-AE) in~\cite{kaufmann2020convolutional}, LEMO~\cite{lemo2021} and PoseNet~\cite{wang2021synthesizing}. We have chosen these baselines as they are the closest ones compared with our setting. For fair comparisons, we use the same body markers and the trajectory representation in all experiments\textsuperscript{\ref{footnote_implementation}}.

\myparagraph{Evaluation Metrics.} \textit{(1) 3D marker accuracy.} For deterministic models, we measure the marker prediction accuracy by computing the Average L2 Distance Error (ADE) between the predicted markers and ground truth. For our generative model, we follow~\cite{yuan2020dlow} to measure the sample accuracy by computing the minimal error between the ground truth and 10 random samples.

\textit{(2) Motion smoothness.} %
We follow~\cite{lemo2021} to use PSKL-J to measure the Power Spectrum KL divergence between the acceleration distribution of synthesized and ground truth joint motion sequences. PSKL-J being non-symmetric, we show the results of both direction, {\emph{i.e.}}, (Predicted, Ground Truth) and (Ground Truth, Predicted). \textit{(3) Foot skating.} Following~\cite{MOJO}, we measure the foot skating artifacts during motion and define skating as when the heel is within 5cm of the ground and the heel speed of both feet exceeds 75mm/s. \textit{(4) foot--ground collision.} We also use a non-collision score, defined as the number of body mesh vertices above the ground divided by the total number of vertices.

\myparagraph{Results.} In Table~\ref{tab:exp-motionfillvae}, we compare our generative motion infilling model with the deterministic Route+PoseNet baseline \cite{wang2021synthesizing}, and both methods can infill the global trajectory and local pose motion. The results show that our generative model can yield much lower average 3D marker distance error (ADE). Also, our method has less foot skating and lower PSKL-J scores in both directions, which demonstrates that our method can generate more natural motions. We also compare our stochastic local motion infilling component (\textbf{Ours}-Local) against other deterministic local motion infilling baselines in Table~\ref{tab:exp-motionfillvae}. Our method outperforms all the other baselines in ADE, foot skating and PSKL-J, demonstrating that the our generative model can better capture human motion patterns and generate more natural motions. 
The motion sequences from the GRAB dataset and our generated motions have non-collision score of 0.9771 and 0.9743, respectively, showing that our method can effectively prevent foot--ground interpenetration.

\vspace{-0.6em}
\subsection{Whole-body Grasp Motion Synthesis}
\label{exp-fillpipeline}
\myparagraph{Experiment setup.} We test our grasping motion generation pipeline on 14 unseen objects from GRAB and HO3D dataset, and we generate 2s motions to grasp the object. Given different initial human poses, we place objects in front of the human at different heights (0.5m--1.7m) with various orientations (0--360$^\circ$ around the gravity axis) and different distances from start point to objects (5cm--1.1m). We conduct user studies on Amazon Mechanical Turk (AMT) for both ground truth grasping motion sequences from GRAB and our generated samples. %
On a five-point scale, three users are asked to rate the realism of presented motions, ranging from \textit{strongly disagree} (score 0) to \textit{strongly agree} (score 5)\textsuperscript{\ref{footnote_implementation}}.

\myparagraph{Results.} The perceptual scores for ground truth sequences and our synthesized sequences are 4.04 (around \textit{agree}) and 3.15 (above \textit{slightly agree}) respectively, showing that our proposed pipeline can synthesize high-fidelity grasping motions. %

\vspace{-0.6em}
\section{Conclusion and Discussion}
\label{sec:conclusion}

In this work, we address an important task on how to synthesize realistic whole-body grasping motion. 
We propose a new approach consisting of two stages: (a) a WholeGrasp-VAE to generate static whole-body grasping poses; (b) a MotionFill-VAE to infill the grasp-oriented motion, given an initial pose and the predicted end pose. Our method, SAGA, is able to generate diverse motion sequences that have realistic interactions with the ground and random objects. We believe SAGA makes progress towards synthesizing human--object interaction, and provides a useful tool for computer graphics and robotics applications. 
However, in this work, we focus on the human motion synthesis task where a virtual human approaches to grasp an object without further hand--object manipulation.  
A future work is to synthesize the hand--object manipulation, while taking the object affordance, physics and the goal of the interaction into account. 

\myparagraph{Acknowledgement.}
This work was supported by the SNF grant 200021 204840 and Microsoft Mixed Reality \& AI Zurich Lab PhD scholarship. We also thank Omid Taheri and Dimitrios Tzionas for helpful discussions.

\clearpage
\bibliographystyle{splncs04}
\bibliography{egbib}

\begin{thebibliography}{10}
\providecommand{\url}[1]{\texttt{#1}}
\providecommand{\urlprefix}{URL }
\providecommand{\doi}[1]{https://doi.org/#1}

\bibitem{alahi2014socially}
Alahi, A., Ramanathan, V., Fei-Fei, L.: Socially-aware large-scale crowd
  forecasting. In: Proceedings of the IEEE Conference on Computer Vision and
  Pattern Recognition. pp. 2203--2210 (2014)

\bibitem{barsoum2018hp}
Barsoum, E., Kender, J., Liu, Z.: Hp-gan: Probabilistic 3d human motion
  prediction via gan. In: Proceedings of the IEEE conference on computer vision
  and pattern recognition workshops. pp. 1418--1427 (2018)

\bibitem{Brahmbhatt_2019_CVPR}
Brahmbhatt, S., Ham, C., Kemp, C.C., Hays, J.: {ContactDB}: Analyzing and
  predicting grasp contact via thermal imaging. In: The IEEE Conference on
  Computer Vision and Pattern Recognition (CVPR) (2019)

\bibitem{brahmbhatt2019contactgrasp}
Brahmbhatt, S., Handa, A., Hays, J., Fox, D.: {ContactGrasp: Functional
  Multi-finger Grasp Synthesis from Contact}. In: 2019 IEEE/RSJ International
  Conference on Intelligent Robots and Systems (IROS) (2019)

\bibitem{cai2020learning}
Cai, Y., Huang, L., Wang, Y., Cham, T.J., Cai, J., Yuan, J., Liu, J., Yang, X.,
  Zhu, Y., Shen, X., et~al.: Learning progressive joint propagation for human
  motion prediction. In: European Conference on Computer Vision. pp. 226--242.
  Springer (2020)

\bibitem{cai2021unified}
Cai, Y., Wang, Y., Zhu, Y., Cham, T.J., Cai, J., Yuan, J., Liu, J., Zheng, C.,
  Yan, S., Ding, H., et~al.: A unified 3d human motion synthesis model via
  conditional variational auto-encoder. In: Proceedings of the IEEE/CVF
  International Conference on Computer Vision. pp. 11645--11655 (2021)

\bibitem{cao2020long}
Cao, Z., Gao, H., Mangalam, K., Cai, Q.Z., Vo, M., Malik, J.: Long-term human
  motion prediction with scene context. In: European Conference on Computer
  Vision. pp. 387--404. Springer (2020)

\bibitem{charbonnier}
Charbonnier, P., Blanc-Feraud, L., Aubert, G., Barlaud, M.: Two deterministic
  half-quadratic regularization algorithms for computed imaging. In:
  Proceedings of 1st International Conference on Image Processing. vol.~2, pp.
  168--172 vol.2 (1994)

\bibitem{chiu2019action}
Chiu, H.k., Adeli, E., Wang, B., Huang, D.A., Niebles, J.C.: Action-agnostic
  human pose forecasting. In: 2019 IEEE Winter Conference on Applications of
  Computer Vision (WACV). pp. 1423--1432. IEEE (2019)

\bibitem{5509126}
Detry, R., Kraft, D., Buch, A.G., Krüger, N., Piater, J.: Refining grasp
  affordance models by experience. In: 2010 IEEE International Conference on
  Robotics and Automation. pp. 2287--2293 (2010)

\bibitem{fragkiadaki2015recurrent}
Fragkiadaki, K., Levine, S., Felsen, P., Malik, J.: Recurrent network models
  for human dynamics. In: Proceedings of the IEEE International Conference on
  Computer Vision. pp. 4346--4354 (2015)

\bibitem{grady2021contactopt}
Grady, P., Tang, C., Twigg, C.D., Vo, M., Brahmbhatt, S., Kemp, C.C.:
  {ContactOpt}: Optimizing contact to improve grasps. In: Conference on
  Computer Vision and Pattern Recognition (CVPR) (2021)

\bibitem{gupta20113d}
Gupta, A., Satkin, S., Efros, A.A., Hebert, M.: From 3d scene geometry to human
  workspace. In: CVPR 2011. pp. 1961--1968. IEEE (2011)

\bibitem{gupta2018social}
Gupta, A., Johnson, J., Fei-Fei, L., Savarese, S., Alahi, A.: Social gan:
  Socially acceptable trajectories with generative adversarial networks. In:
  Proceedings of the IEEE Conference on Computer Vision and Pattern
  Recognition. pp. 2255--2264 (2018)

\bibitem{hampali2020honnotate}
Hampali, S., Rad, M., Oberweger, M., Lepetit, V.: Honnotate: A method for 3d
  annotation of hand and object poses. In: Proceedings of the IEEE/CVF
  conference on computer vision and pattern recognition. pp. 3196--3206 (2020)

\bibitem{harvey2020robust}
Harvey, F.G., Yurick, M., Nowrouzezahrai, D., Pal, C.: Robust motion
  in-betweening. ACM Transactions on Graphics (TOG)  \textbf{39}(4),  60--1
  (2020)

\bibitem{helbing1995social}
Helbing, D., Molnar, P.: Social force model for pedestrian dynamics. Physical
  review E  \textbf{51}(5), ~4282 (1995)

\bibitem{hernandez2019human}
Hernandez, A., Gall, J., Moreno-Noguer, F.: Human motion prediction via
  spatio-temporal inpainting. In: Proceedings of the IEEE/CVF International
  Conference on Computer Vision. pp. 7134--7143 (2019)

\bibitem{holden2017phase}
Holden, D., Komura, T., Saito, J.: Phase-functioned neural networks for
  character control. ACM Transactions on Graphics (TOG)  \textbf{36}(4),  1--13
  (2017)

\bibitem{holden2016deep}
Holden, D., Saito, J., Komura, T.: A deep learning framework for character
  motion synthesis and editing. ACM Transactions on Graphics (TOG)
  \textbf{35}(4),  1--11 (2016)

\bibitem{hsiao2006imitation}
Hsiao, K., Lozano-Perez, T.: Imitation learning of whole-body grasps. In: 2006
  IEEE/RSJ international conference on intelligent robots and systems. pp.
  5657--5662. IEEE (2006)

\bibitem{jain2016structural}
Jain, A., Zamir, A.R., Savarese, S., Saxena, A.: Structural-rnn: Deep learning
  on spatio-temporal graphs. In: Proceedings of the ieee conference on computer
  vision and pattern recognition. pp. 5308--5317 (2016)

\bibitem{jiang2021graspTTA}
Jiang, H., Liu, S., Wang, J., Wang, X.: Hand-object contact consistency
  reasoning for human grasps generation. In: Proceedings of the International
  Conference on Computer Vision (2021)

\bibitem{kalisiak2001grasp}
Kalisiak, M., Van~de Panne, M.: A grasp-based motion planning algorithm for
  character animation. The Journal of Visualization and Computer Animation
  \textbf{12}(3),  117--129 (2001)

\bibitem{GrapingField:3DV:2020}
Karunratanakul, K., Yang, J., Zhang, Y., Black, M., Muandet, K., Tang, S.:
  Grasping field: Learning implicit representations for human grasps. In: 8th
  International Conference on 3D Vision. pp. 333--344. {IEEE} (Nov 2020)

\bibitem{kaufmann2020convolutional}
Kaufmann, M., Aksan, E., Song, J., Pece, F., Ziegler, R., Hilliges, O.:
  Convolutional autoencoders for human motion infilling. In: 2020 International
  Conference on 3D Vision (3DV). pp. 918--927. IEEE (2020)

\bibitem{kingma2014adam}
Kingma, D.P., Ba, J.: Adam: A method for stochastic optimization. arXiv
  preprint arXiv:1412.6980  (2014)

\bibitem{5654380}
Krug, R., Dimitrov, D., Charusta, K., Iliev, B.: On the efficient computation
  of independent contact regions for force closure grasps. In: 2010 IEEE/RSJ
  International Conference on Intelligent Robots and Systems. pp. 586--591
  (2010)

\bibitem{10.1145/1141911.1141969}
Kry, P.G., Pai, D.K.: Interaction capture and synthesis. ACM Trans. Graph.
  \textbf{25}(3),  872–880 (Jul 2006)

\bibitem{li2021task}
Li, J., Villegas, R., Ceylan, D., Yang, J., Kuang, Z., Li, H., Zhao, Y.:
  Task-generic hierarchical human motion prior using vaes. In: 2021
  International Conference on 3D Vision (3DV). pp. 771--781. IEEE (2021)

\bibitem{li2019putting}
Li, X., Liu, S., Kim, K., Wang, X., Yang, M.H., Kautz, J.: Putting humans in a
  scene: Learning affordance in 3d indoor environments. In: Proceedings of the
  IEEE Conference on Computer Vision and Pattern Recognition. pp. 12368--12376
  (2019)

\bibitem{4293017}
Li, Y., Fu, J.L., Pollard, N.S.: Data-driven grasp synthesis using shape
  matching and task-based pruning. IEEE Transactions on Visualization and
  Computer Graphics  \textbf{13}(4),  732--747 (2007)

\bibitem{ling2020character}
Ling, H.Y., Zinno, F., Cheng, G., Van De~Panne, M.: Character controllers using
  motion vaes. ACM Transactions on Graphics (TOG)  \textbf{39}(4),  40--1
  (2020)

\bibitem{liu2018learning}
Liu, L., Hodgins, J.: Learning basketball dribbling skills using trajectory
  optimization and deep reinforcement learning. ACM Transactions on Graphics
  (TOG)  \textbf{37}(4),  1--14 (2018)

\bibitem{liu2019generating}
Liu, M., Pan, Z., Xu, K., Ganguly, K., Manocha, D.: Generating grasp poses for
  a high-dof gripper using neural networks. In: 2019 IEEE/RSJ International
  Conference on Intelligent Robots and Systems (IROS). pp. 1518--1525. IEEE
  (2019)

\bibitem{Lucas2019UnderstandingPC}
Lucas, J., Tucker, G., Grosse, R.B., Norouzi, M.: Understanding posterior
  collapse in generative latent variable models. In: DGS@ICLR (2019)

\bibitem{AMASS:ICCV:2019}
Mahmood, N., Ghorbani, N., Troje, N.F., Pons-Moll, G., Black, M.J.: {AMASS}:
  Archive of motion capture as surface shapes. In: International Conference on
  Computer Vision. pp. 5442--5451 (Oct 2019)

\bibitem{makansi2019overcoming}
Makansi, O., Ilg, E., Cicek, O., Brox, T.: Overcoming limitations of mixture
  density networks: A sampling and fitting framework for multimodal future
  prediction. In: Proceedings of the IEEE Conference on Computer Vision and
  Pattern Recognition. pp. 7144--7153 (2019)

\bibitem{mao2019learning}
Mao, W., Liu, M., Salzmann, M., Li, H.: Learning trajectory dependencies for
  human motion prediction. In: Proceedings of the IEEE International Conference
  on Computer Vision. pp. 9489--9497 (2019)

\bibitem{martinez2017human}
Martinez, J., Black, M.J., Romero, J.: On human motion prediction using
  recurrent neural networks. In: Proceedings of the IEEE Conference on Computer
  Vision and Pattern Recognition. pp. 2891--2900 (2017)

\bibitem{NEURIPS2019_9015}
Paszke, A., Gross, S., Massa, F., Lerer, A., Bradbury, J., Chanan, G., Killeen,
  T., Lin, Z., Gimelshein, N., Antiga, L., Desmaison, A., Kopf, A., Yang, E.,
  DeVito, Z., Raison, M., Tejani, A., Chilamkurthy, S., Steiner, B., Fang, L.,
  Bai, J., Chintala, S.: Pytorch: An imperative style, high-performance deep
  learning library. In: Wallach, H., Larochelle, H., Beygelzimer, A.,
  d\textquotesingle Alch\'{e}-Buc, F., Fox, E., Garnett, R. (eds.) Advances in
  Neural Information Processing Systems 32, pp. 8024--8035. Curran Associates,
  Inc. (2019)

\bibitem{SMPL-X:2019}
Pavlakos, G., Choutas, V., Ghorbani, N., Bolkart, T., Osman, A.A.A., Tzionas,
  D., Black, M.J.: Expressive body capture: {3D} hands, face, and body from a
  single image. In: Proceedings IEEE Conf. on Computer Vision and Pattern
  Recognition (CVPR). pp. 10975--10985 (2019)

\bibitem{pollard2005physically}
Pollard, N.S., Zordan, V.B.: Physically based grasping control from example.
  In: Proceedings of the 2005 ACM SIGGRAPH/Eurographics symposium on Computer
  animation. pp. 311--318 (2005)

\bibitem{BPS}
Prokudin, S., Lassner, C., Romero, J.: Efficient learning on point clouds with
  basis point sets. In: Proceedings of the IEEE/CVF International Conference on
  Computer Vision (ICCV) (October 2019)

\bibitem{qi2017pointnet++}
Qi, C.R., Yi, L., Su, H., Guibas, L.J.: Pointnet++: Deep hierarchical feature
  learning on point sets in a metric space. Advances in Neural Information
  Processing Systems  \textbf{30} (2017)

\bibitem{rempe2021humor}
Rempe, D., Birdal, T., Hertzmann, A., Yang, J., Sridhar, S., Guibas, L.J.:
  Humor: 3d human motion model for robust pose estimation. In: Proceedings of
  the IEEE/CVF International Conference on Computer Vision. pp. 11488--11499
  (2021)

\bibitem{rijpkema1991computer}
Rijpkema, H., Girard, M.: Computer animation of knowledge-based human grasping.
  ACM Siggraph Computer Graphics  \textbf{25}(4),  339--348 (1991)

\bibitem{MANO:SIGGRAPHASIA:2017}
Romero, J., Tzionas, D., Black, M.J.: Embodied hands: Modeling and capturing
  hands and bodies together. ACM Transactions on Graphics, (Proc. SIGGRAPH
  Asia)  \textbf{36}(6) (Nov 2017)

\bibitem{sadeghian2019sophie}
Sadeghian, A., Kosaraju, V., Sadeghian, A., Hirose, N., Rezatofighi, H.,
  Savarese, S.: Sophie: An attentive gan for predicting paths compliant to
  social and physical constraints. In: Proceedings of the IEEE Conference on
  Computer Vision and Pattern Recognition. pp. 1349--1358 (2019)

\bibitem{savva2016pigraphs}
Savva, M., Chang, A.X., Hanrahan, P., Fisher, M., Nie{\ss}ner, M.: Pigraphs:
  learning interaction snapshots from observations. ACM Transactions on
  Graphics (TOG)  \textbf{35}(4),  1--12 (2016)

\bibitem{6225086}
Seo, J., Kim, S., Kumar, V.: Planar, bimanual, whole-arm grasping. In: 2012
  IEEE International Conference on Robotics and Automation. pp. 3271--3277
  (2012)

\bibitem{starke2019neural}
Starke, S., Zhang, H., Komura, T., Saito, J.: Neural state machine for
  character-scene interactions. ACM Trans. Graph.  \textbf{38}(6),  209--1
  (2019)

\bibitem{starke2020local}
Starke, S., Zhao, Y., Komura, T., Zaman, K.: Local motion phases for learning
  multi-contact character movements. ACM Transactions on Graphics (TOG)
  \textbf{39}(4),  54--1 (2020)

\bibitem{GOAL}
Taheri, O., Choutas, V., Black, M.J., Tzionas, D.: Goal: Generating 4d
  whole-body motion for hand-object grasping. arXiv preprint arXiv:2112.11454
  (2021)

\bibitem{GRAB:2020}
Taheri, O., Ghorbani, N., Black, M.J., Tzionas, D.: {GRAB}: A dataset of
  whole-body human grasping of objects. In: European Conference on Computer
  Vision (ECCV) (2020)

\bibitem{tai2018socially}
Tai, L., Zhang, J., Liu, M., Burgard, W.: Socially compliant navigation through
  raw depth inputs with generative adversarial imitation learning. In: 2018
  IEEE International Conference on Robotics and Automation (ICRA). pp.
  1111--1117. IEEE (2018)

\bibitem{tan2018and}
Tan, F., Bernier, C., Cohen, B., Ordonez, V., Barnes, C.: Where and who?
  automatic semantic-aware person composition. In: 2018 IEEE Winter Conference
  on Applications of Computer Vision (WACV). pp. 1519--1528. IEEE (2018)

\bibitem{wang2019imitation}
Wang, B., Adeli, E., Chiu, H.k., Huang, D.A., Niebles, J.C.: Imitation learning
  for human pose prediction. In: Proceedings of the IEEE International
  Conference on Computer Vision. pp. 7124--7133 (2019)

\bibitem{wang2021synthesizing}
Wang, J., Xu, H., Xu, J., Liu, S., Wang, X.: Synthesizing long-term 3d human
  motion and interaction in 3d scenes. In: Proceedings of the IEEE/CVF
  Conference on Computer Vision and Pattern Recognition. pp. 9401--9411 (2021)

\bibitem{yan2019convolutional}
Yan, S., Li, Z., Xiong, Y., Yan, H., Lin, D.: Convolutional sequence generation
  for skeleton-based action synthesis. In: Proceedings of the IEEE/CVF
  International Conference on Computer Vision. pp. 4394--4402 (2019)

\bibitem{yan2018mt}
Yan, X., Rastogi, A., Villegas, R., Sunkavalli, K., Shechtman, E., Hadap, S.,
  Yumer, E., Lee, H.: Mt-vae: Learning motion transformations to generate
  multimodal human dynamics. In: Proceedings of the European Conference on
  Computer Vision (ECCV). pp. 265--281 (2018)

\bibitem{yuan2020dlow}
Yuan, Y., Kitani, K.: Dlow: Diversifying latent flows for diverse human motion
  prediction. In: Proceedings of the European Conference on Computer Vision
  (ECCV) (2020)

\bibitem{Zhang2021ManipNetNM}
Zhang, H., Ye, Y., Shiratori, T., Komura, T.: Manipnet: neural manipulation
  synthesis with a hand-object spatial representation. ACM Trans. Graph.
  \textbf{40},  121:1--121:14 (2021)

\bibitem{lemo2021}
Zhang, S., Zhang, Y., Bogo, F., Pollefeys, M., Tang, S.: Learning motion priors
  for 4d human body capture in 3d scenes. In: IEEE/CVF International Conference
  on Computer Vision (ICCV 2021) (2021)

\bibitem{MOJO}
Zhang, Y., Black, M.J., Tang, S.: We are more than our joints: Predicting how
  3d bodies move. In: Proceedings of the IEEE/CVF Conference on Computer Vision
  and Pattern Recognition. pp. 3372--3382 (2021)

\bibitem{zhang2020learning}
Zhang, Y., Yu, W., Liu, C.K., Kemp, C., Turk, G.: Learning to manipulate
  amorphous materials. ACM Transactions on Graphics (TOG)  \textbf{39}(6),
  1--11 (2020)

\bibitem{Zhou_2019_CVPR}
Zhou, Y., Barnes, C., Lu, J., Yang, J., Li, H.: On the continuity of rotation
  representations in neural networks. In: Proceedings of the IEEE/CVF
  Conference on Computer Vision and Pattern Recognition (CVPR) (June 2019)

\end{thebibliography}

\newpage
\renewcommand{\thefigure}{S\arabic{figure}}
\renewcommand{\thetable}{S\arabic{table}}
\renewcommand{\theequation}{S.\arabic{equation}}
\title{SAGA: Stochastic Whole-Body Grasping with Contact\\ ** Appendix **} %
\titlerunning{SAGA-Appendix}
\author{Yan Wu$^*$\inst{1},
        Jiahao Wang$^*$\inst{2},
        Yan Zhang\inst{1},
        Siwei Zhang\inst{1},
        Otmar Hilliges\inst{1},
        Fisher Yu\inst{1},
        Siyu Tang\inst{1}}
\authorrunning{Y. Wu$^*$, J. Wang$^*$ et al.}
\institute{ETH Z\"urich, Switzerland \and Max Planck Institute for Informatics, Germany\\
\email{yan.wu@vision.ee.ethz.ch, jiwang@mpi-inf.mpg.de, \{yan.zhang,siwei.zhang,otmar.hilliges,siyu.tang\}@inf.ethz.ch, i@yf.io}}
\maketitle
\def\thefootnote{*}\footnotetext{Equal contribution.}

\appendix
\setcounter{figure}{0} 
\setcounter{table}{0} 
\setcounter{equation}{0} 
In the appendix, we first provide the body markers representation, architecture details, training and optimization setup, dataset pre-processing details, and the AMT user study evaluation details in Appendix~\ref{appen-Implementation-details}. In Appendix~\ref{appen-baselines}, we illustrate the detailed baseline experiments and ablation study setup. We further provide additional experimental results and visualization results in Appendix~\ref{appen-exp}, and we discuss the existing limitations in Appendix~\ref{appen-limitations}. Please see the \href{https://jiahaoplus.github.io/SAGA/saga.html}{\textbf{video}} at our project page for more random samples of synthesized grasping poses and grasping motions.

\section{Method and Implementation Details}
\label{appen-Implementation-details}
\subsection{Body Markers Placement}
\begin{figure}
    \centering
    \includegraphics[width=1\linewidth]{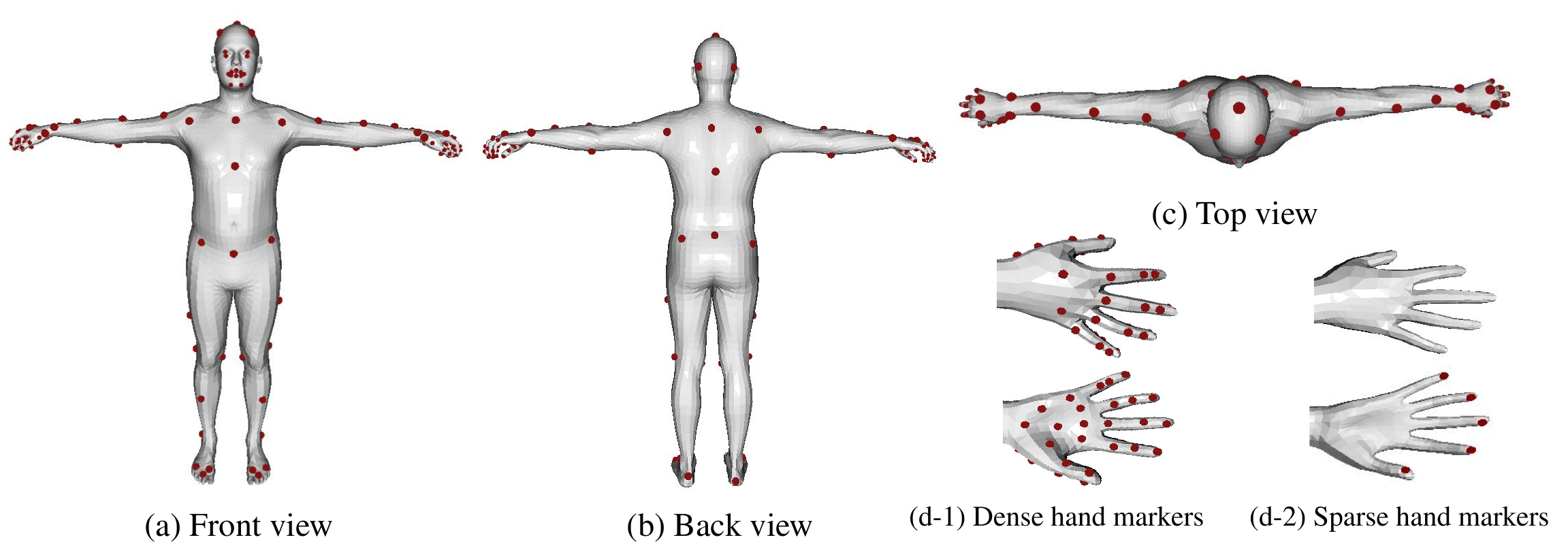}
    \caption{Visualization of our body markers placement. We have 69 markers on the body surface, which are visualized as red spheres on SMPL-X body surface, in which 49 for the body, 14 for the face and 6 for hands. For the WholeGrasp-VAE training, we additionally have 30 markers on fingers and 44 markers on palms (d-1). For the MotionFill-VAE training, we only have sparse hand markers (d-2) with 10 markers on fingertips.}
    \label{fig:markers-setup}
\end{figure}

To have informative yet compact markers setup, as illustrated in Fig.~\ref{fig:markers-setup}, we follow the placement of the markers in GRAB~\cite{GRAB:2020} MoCap system, having 49 markers for the body, 14 for the face, 6 for hands and 30 for fingers (see Fig.~\ref{fig:markers-setup} (a)-(c)) on SMPL-X %
body surface. As hand poses are subtle and the palm is frequently in contact with the object, we additionally have 44 markers on two palms (see Fig~\ref{fig:markers-setup} (d-1)) to further enrich the markers information for a better grasp. For the grasping ending pose generation in stage1, we use all these 143 markers for training and optimization. For the motion infilling network training in stage2, we only use a sparse set of palm markers with 10 markers on fingertips (see Fig.~\ref{fig:markers-setup}(d-2)).
\subsection{Architecture Details}
\subsubsection{WholeGrasp-VAE}
We have visualized the WholeGrasp-VAE architecture in Fig.~\ref{fig:Arch-WholeGraspCVAE}. This CVAE is conditioned on the object height information and the object geometry feature extracted with PointNet++~\cite{qi2017pointnet++} encoder. In the encoder, taking the body markers' postions $\bm{M} \in \mathbb{R}^{N\times3}$ and body markers contacts $C_{\bm{M}} \in \{0, 1\}^{N}$ as inputs, where N is the number of markers, the body branch encodes the body feature $\mathcal{F}_B$; Taking the object contacts $C_{\bm{O}} \in \{0, 1\}^{2048}$ as an additional feature of the object point cloud data, the object branch uses the PointNet++ to encode the object feature. Further, we fuse them into a joint 16-dimensional latent space $\bm{z}_{s}$. In the decoder, we individually decode the body markers' positions, markers' contacts, and object contacts. Note that we model the contacts learning as a two-class (\textit{in-contact} or \textit{out-of-contact}) classification task, and the decoder outputs the \textit{in-contact} probability of each points. And the PointNet++ encoder architecture is given by: SA(256, 0.2, [64, 128]) $\rightarrow$ SA(128, 0.25, [128, 256]) $\rightarrow$ SA([256, 512]), where SA denotes the set abstraction level~\cite{qi2017pointnet++}. 
\subsubsection{MotionFill-VAE}
In TrajFill, the root state at time $t$ is given by $\bm{\Gamma}_t = (x_t, y_t,$ $ \cos\gamma_t, \sin\gamma_t)$, where $x_t, y_t$ are the position of pelvis joint in the x-y (ground) plane, and $\gamma_t$ is the body rotation around z-axis. Given $\bm{\Gamma}_{0}$ and $\bm{\Gamma}_{T}$, the TrajFill is built to learn the deviation $\Delta \bm{\Gamma}_{0:T+1} = \bm{\Gamma}_{0:T+1} - \overline{\bm{\Gamma}}_{0:T+1}$ from an initial trajectory $\overline{\bm{\Gamma}}_{0:T+1}$ which is a linear interpolation and one-step extrapolation of the given $\bm{\Gamma}_0$ and $\bm{\Gamma}_T$, and we use $\overline{\bm{\Gamma}}_{0:T+1}$ as the condition. Inside TrajFill, we use MLP structures for the encoder and the decoder. For the encoder, input trajectory features are passed through two residual blocks, which has the same hidden size as the input dimension (8$T$, where $T$ is the time length of the input). After that, two linear branches project the features into the 512-dimensional latent space. The decoder includes two residual blocks with hidden sizes equal to 8$T$ and 4$T$, respectively. We get the final output of TrajFill by adding the last residual block output and the initial rough trajectory $\overline{\bm{\Gamma}}_{0:T+1}$.

In LocalMotionFill, following the same input processing step as in~\cite{lemo2021}, we build a 4-channel local motion image $\bm{I} \in \mathbb{R}^{4\times(3N+n)\times T}$, where $N, n$ are the number of markers and the dimension of foot-ground contact labels. The first channel of $\bm{I}$ is a concatenation of foot-ground contacts $C_{F_{0:T}} \in \{0, 1\}^{n \times T}$ and the normalized local markers $\bm{M}^l_{0:T} \in \mathbb{R}^{3N \times T}$.%
The other three channels of $\bm{I}_l$ are the normalized root local velocities $\bm{v}_{0:T}^l$. To incorporate the condition information ($\bm{M}_0, \bm{M}_T, \bm{v}_{0:T}$) into the LocalMotionFill, similarly, we build a condition image $\bm{I}_c$ which essentially is the masked input image $\bm{I}$. We use the same CNN-based encoder and decoder network as in \cite{lemo2021} to learn the infilled motion image.

\subsection{Dataset Processing}
\myparagraph{GRAB.}
We use GRAB (https://grab.is.tue.mpg.de/license) dataset to train both the WholeGrasp-VAE and MotionFill-VAE for grasping ending pose generation and motion infilling, respectively. 

For WholeGrasp-VAE training, considering the different body shape pattern of male and female, we suggest training the male model and the female model separately. Following GrabNet~\cite{GRAB:2020}, for training, we take all frames with right-hand grasps. And out of the 51 different objects in GRAB dataset, following the same split of object class in GrabNet\cite{GRAB:2020, Brahmbhatt_2019_CVPR}  we take out 4 validation objects (\textit{apple, toothbrush, elephant, hand}) and 6 test objects (\textit{mug, camera, toothpaste, wineglass, frying pan, binoculars}), and the remaining 41 objects are used for training. We center the object point cloud and the body markers at the geometry center of the object.

For MotionFill-VAE training, we only utilize sequences where humans are approaching to grasp the object. GRAB dataset captures the motion sequences where the human starts with T-pose, approaches and grasp the object, and then interacts with the object. For MotionFill-VAE training, we clip those approaching and grasping sequences. Since most of these approaching motion sequences only last for about 2s in the GRAB dataset, we clip 2-second videos from each sequence by ensuring that the last frames are at stable grasping poses. If the sequence is shorter than 2s, we pad the first frame to have the two-second clip.

\myparagraph{AMASS.}
We pre-train our motion infilling model MotionFill-VAE on the AMASS (https://amass.is.tue.mpg.de/license.html). We down-sample the sequences to 30 fps and cut them into clips with same duration. To be consistent with GRAB dataset, we clip 2-second sequences from the AMASS for the grasping motion infilling task. We also evaluate our motion infilling network by conducting experiments on the general motion infilling task with different time lengths (see Appendix~\ref{appen-exp-motioninfill}), and for that, we clip the AMASS dataset into 4-second and 6-second sequences. Similar to \cite{MOJO}, we reset the world coordinate for each clip. The origin of the world coordinate is set to the pelvis joint in the first frame. The x-axis is the horizontal component of the direction from the left shoulder to right shoulder, the y-axis faces forward, and the z-axis points upward. For training, we use all the mocap datasets except EKUT, KIT, SFU, SSM synced, TCD handMocap, and TotalCapture. For testing, we use TCD handMocap, TotalCapture, and SFU. %

\begin{minipage}{0.8\textwidth}
\begin{algorithm}[H]
\caption{\small{WholeGrasp-Opt: grasping pose optimization}}\label{alg:WholeGrasp-Opt}
\small{
\textbf{Input:} Sampled markers $\hat{\bm{M}}$, markers-object contacts $\hat{C}_{\bm{M}}$, $\hat{C}_{\bm{O}}$.\\
\textbf{Output:} Body mesh $\bm{B}_T(\bm{\Theta}_T)$ and the queried markers $\bm{M}_T$.\\
\textbf{Require:} Optimization steps $(N_1, N_2, N_3)=(300, 400, 500)$.
}
\begin{algorithmic}
\For{i = $1:N_1$}
    \State {Optimize $\bm{t}, \bm{R}$ to minimize $E_{fit}$ in Eq.~\ref{eq:opt-1-fit}}.
\EndFor
\For{i = $N_1:N_1+N_2$}
    \State {Optimize $\bm{t}, \bm{R}, \bm{\beta}, \bm{\theta}_b$ to minimize $E_{fit}$ in Eq.~\ref{eq:opt-1-fit}}.
\EndFor
\For{i = $N_1+N_2:N_1+N_2+N_3$}
    \State {Optimize $\bm{\theta}_b, \bm{\theta}_h, \bm{\theta}_e$ to minimize $E_{opt}$ in Eq.~\ref{eq:opt-1}}-\ref{eq:opt-1-contact}.
\EndFor
\State \Return {$\bm{\Theta}_T=[\bm{\beta, t, R}, \bm{\theta}_b, \bm{\theta}_h, \bm{\theta}_e]$}
\end{algorithmic}
\end{algorithm}
\end{minipage}
\subsection{Implementation Details}
We implement our experiments using PyTorch v1.6.0~\cite{NEURIPS2019_9015}. In the following, we introduce the training details of WholeGrasp-VAE and MotionFill-VAE, and optimization details of GraspPose-Opt and GraspMotion-Opt respectively.

\myparagraph{WholeGrasp-VAE training.} In Sec.~\ref{WholeGrasp}, we have introduced the WholeGrasp-VAE training losses. Note that for the object and markers contact map reconstruction, due to the class in-balance, we employ the weighted binary cross-entropy loss, and we empirically set the weights for \textit{in-contact} class for objects and markers as 3 and 5 respectively. And we set the object and markers contact map reconstruction weight $\lambda_{\bm{M}}, \lambda_{\bm{O}}$ in Eq.~\ref{eq:train-1-rec-loss} as 1. For the VAE training, we adopt the linear KL weight annealing strategy~\cite{Lucas2019UnderstandingPC} to avoid posterior collapse issue in VAE training. And we empirically set $\lambda_c=1$ and $\lambda_{KL}=0.005e$ in $\mathcal{L}_{train} = \mathcal{L}_{rec} + \lambda_{KL}\mathcal{L}_{KL} + \lambda_c \mathcal{L}_{c}$, where $e$ is the epoch number, and we train the WholeGrasp-VAE for 40 epochs.

\myparagraph{MotionFill-VAE training.} In the experiments for Table.~\ref{tab:exp-motionfillvae}, for local motion infilling (given the starting pose, ending pose and trajectory), we train our LocalMotionFill model on AMASS training set; for the entire MotionFill-VAE (``Traj + local motion infilling"), we first pretrain our TrajFill module and LocalMotionFill module on the GRAB and AMASS training set respectively, and we further finetune the entire MotionFill-VAE on the GRAB training set. We empirically set the hyper-parameters in $\mathcal{L}_M = \mathcal{L}_{rec}+\lambda_{KL} \mathcal{L}_{KL}$ and Eq.~\ref{eq:motionfill-rec} as follows: \{$\lambda_{KL}, \lambda_{1}, \lambda_{2}, \lambda_{3}, \lambda_{4}$\} = \{1, 0.05, 1, 1, 0.5\}.

\myparagraph{GraspPose-Opt optimization.} In Sec.~\ref{WholeGrasp}, we have illustrated the GraspPose-Opt optimization losses design to recover SMPL-X body mesh from sparse markers and refine the body pose for more perceptually realistic human-object interactions. We empirically set the hyper-parameters in Eq.~\ref{eq:opt-1-fit}-\ref{eq:opt-1-colli}  \{$\alpha_{cont}^o$, $\alpha_{cont}^m$, $\alpha_{colli}^{\bm{O}}$, $\alpha_{colli}^{\bm{B}}$, $\alpha_{\bm{\theta}}$\} = \{15, 15, 100, 200, 0.0005\}. As it can be difficult for the optimization process to converge by jointly optimizing the high-dimensional SMPL-X parameters, which include the body global configuration $\bm{t}$ and $\bm{R}$, shape parameters $\bm{\beta}$, body pose parameters $\bm{\theta}_b$, and the more local hand pose $\bm{\theta}_h$ and eye pose $\bm{\theta}_e$ parameters, similar as in MOJO~\cite{MOJO}, we suggest a multi-stage optimization mechanism by optimizing in a \textit{global-to-local} fashion to facilitate a gradual convergence. And the detailed multi-stage training process can be found in Alg.~\ref{alg:WholeGrasp-Opt}. We use Adam~\cite{kingma2014adam} optimizer, and the initial learning rates for these three stages are set as 0.016, 0.012, 0.008 respectively.

\myparagraph{GraspMotion-Opt optimization.} In Sec.~\ref{fullpipeline}, we have shown the GraspMotion-Opt optimization losses design to recover smooth SMPL-X body motions from sparse markers sequence. Additionally, we introduce more details about our motion smoothness loss and foot skating loss design.
\begin{itemize}
    \item \textbf{Cross-frame smoothness loss.} To encourage a temporarily smooth whole-body motion, following~\cite{lemo2021}, we enforce smoothness on the smooth motion latent space \scalebox{0.9}{$S_{1:T-1}=AE(\bm{M}_{1:T}-\bm{M}_{0:T-1})$} encoded by a pretrained autoencoder. Also, we explicitly enforce smoothness on the hand vertices, and the overall smoothness loss is given by:
\begin{align}
    E_{smooth} = \alpha_s^{B}\sum_{t=1}^{T-2}|S_{t+1}-S_{t}|^2 + \alpha_s^{h}\sum_{t=0}^{T-1}|\mathcal{V}^h_{{\bm{B}}_{t+1}} - \mathcal{V}^h_{{\bm{B}}_{t}}|^2
\end{align}
\item \textbf{Foot skating loss.}  Following~\cite{lemo2021}, we reduce the foot skating artifacts by optimization based on the foot-ground contact labels $\hat{C}_F$. 
\begin{align}
E_{skat}=\alpha_{skat}\sum_{t \in T_c}\sum_{|v_t^{foot}| \geq \sigma}||v_t^{foot}| - \sigma|
\end{align}
where $T_c$ means the timestamps with foot-ground contact, $v_t^{foot}$ represents the velocity (location difference between adjacent timestamps $t$ and $t+1$) of vertices on the left toe, left heel, right toe, and right heel, at time $t$. $\sigma$ is a threshold and we use $\sigma=0.1$ in our experiments.
\end{itemize}

The overall optimization loss is given by $E_{basic}+E_{g}+E_{smooth}+E_{skat}$, where $E_{basic}$ and $E_{g}$ are formulated in Eq.~\ref{eq:opt-2-basic} - \ref{eq:opt-2-grasp}. We optimize the overall loss in two stages, where we first fit SMPL-X body mesh to the predicted markers sequences by minimizing the markers fitting loss ($E_{fit}$ in Eq.~\ref{eq:opt-2-basic}), and then we refine the recovered body mesh sequences by minimizing the overall loss. We present the detailed optimization procedure in Alg.~\ref{alg:GraspMotion-Opt}. We also use Adam\cite{kingma2014adam} optimizer. For the first frame, the initial learning rate is 0.1, and for the other frames the initial learning rate is 0.01. Stage 1 optimization takes 100 steps and the learning rate becomes 0.01 after step 60 and decreases to 0.003 after step 80. The second stage optimization takes 300 steps, and the initial learning rate is set to 0.01 and decays to 0.005 after 150 steps.

\begin{algorithm}[H]
\caption{\small{GraspMotion-Opt: grasping motion optimization}}\label{alg:GraspMotion-Opt}
\small{
\textbf{Input:} Sampled body markers sequences $\hat{\bm{M}}_{0:T}$, mutual markers-object contacts $\hat{C}_{\bm{M}}$, $\hat{C}_{\bm{O}}$ and dynamic foot-ground contact $\hat{C}_{F_{0:T}}$; Body shape parameters $\bm{\beta}$\\
\textbf{Output:} Smoothed whole-body grasping motion $\bm{B}_{0:T}(\bm{\Theta}_{0:T})$.\\
\textbf{Require:} Optimization steps $(N_1, N_2)=(100, 300)$.
}
\begin{algorithmic}

\For{i = $1:N_1$}
    \State {Optimize $[\bm{t, R, \theta}]_{0:T}$ to minimize \scalebox{0.9}{$\sum\limits_{t=0}^TE_{fit}$} in Eq.~\ref{eq:opt-2-basic}}.
\EndFor
\For{i = $N_1:N_1+N_2$}
    \State{Minimize $E_{basic}+E_{g}+E_{smooth}+E_{skat}$ in \S~\ref{fullpipeline} %
    }
\EndFor
\State \Return $\bm{\Theta}_{0:T}= \{\bm{\beta}, [\bm{t, R, \theta}]_{0:T}\}$
\end{algorithmic}
\end{algorithm}

\subsection{Amazon Mechanical Turk (AMT) User Study}
\begin{figure}
    \centering
    \includegraphics[width=\linewidth]{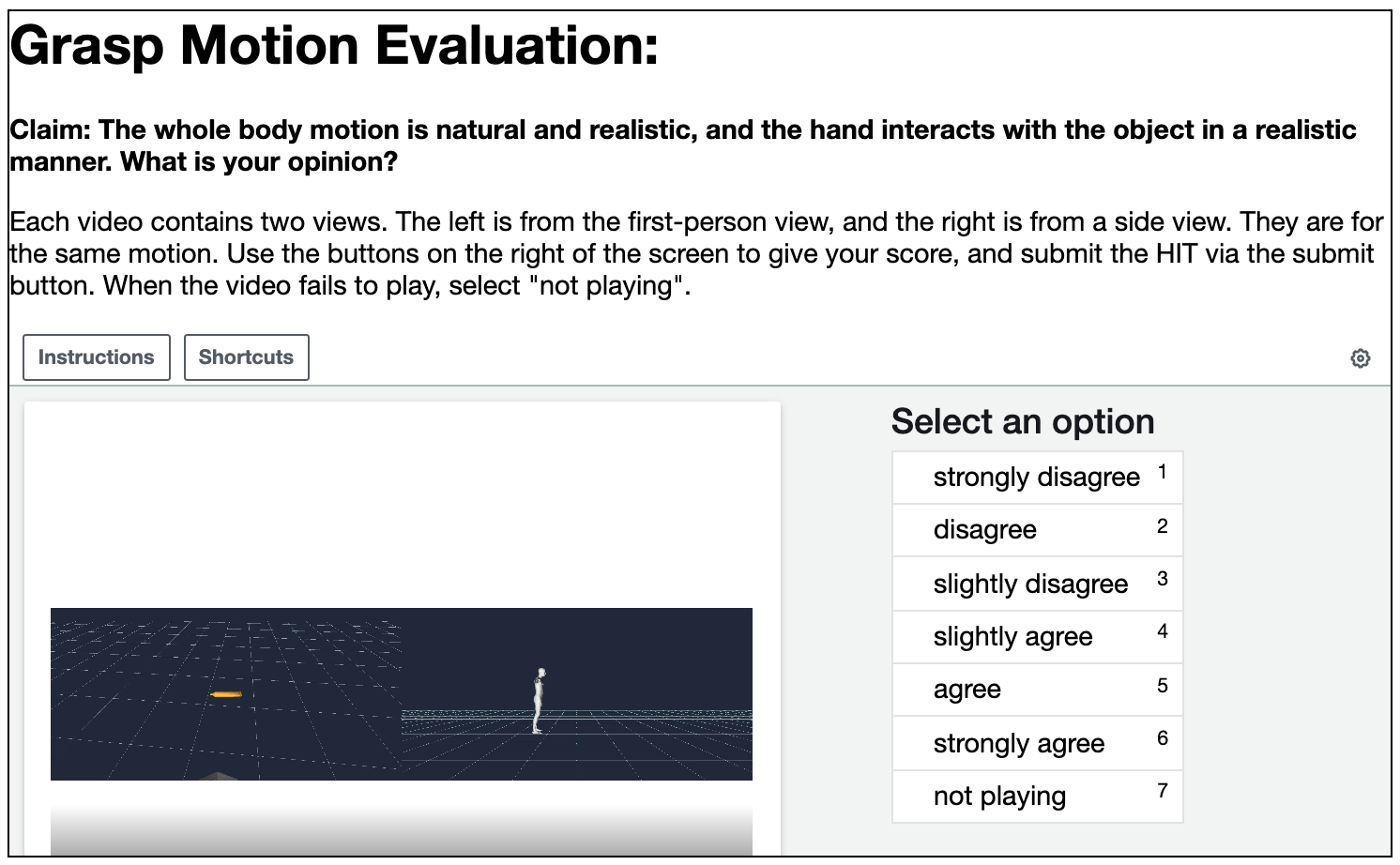}
    \caption{AMT user study interface. We present both the first-view video (on the left side) and third-view video (on the right side) to the user for the grasping motion quality evaluation.}
    \label{fig:AMT_interface}
\end{figure}

We perform user study via AMT, and the user study interface is presented in Fig.~\ref{fig:AMT_interface}. We perform user study on both ground truth motion sequences from GRAB dataset and our randomly generated sample sequences. We test our pipeline with 14 unseen objects from both GRAB test set and HO3D~\cite{hampali2020honnotate} dataset, and we generate 50 random grasping motion sequences for each object, where objects are randomly placed. Each sequence is scored by 3 users and we take the average score, and the score range from 0 to 5 (from \textit{strongly disagree} to \textit{strongly agree}). The average perceptual score for each object class is presented in Table~\ref{tab:AMT_study}.

\begin{table}[]
    \centering
    \caption{Perceptual score results of both ground truth (GT) grasping motion sequences and our synthesized sequences for grasping various unseen objects. Due to the lack of ground truth whole-body grasping motions for objects in HO3D dataset, we only evaluate ground truth sequences for objects from GRAB dataset.}
    \begin{tabular}{ccccccccc}
    \toprule
    \multicolumn{3}{c}{GRAB} & & \multicolumn{2}{c}{HO3D} && \multicolumn{2}{c}{Average Score}
    \\
    \cline{0-2}
    \cline{5-6}
    \cline{8-9}
     \multirow{2}{*}{Object} & \multicolumn{2}{c}{Score} & & \multirow{2}{*}{Object} & \multirow{2}{*}{Score (Ours)} && \multirow{2}{*}{GT} & \multirow{2}{*}{Ours}\\
     \cline{2-3}
     & GT & Ours & &&  \\
     \toprule
     &&&& Cracker box & 3.15  && \multirow{8}{*}{4.04} & \multirow{8}{*}{3.15}\\
     Binoculars & 3.83 & 2.98 & & Sugar box & 3.45 \\
     Camera & 3.92 & 3.60 && Mustard bottle & 3.49 \\
     Toothpaste & 4.22 & 3.47 && Meat can & 3.43 \\
     Mug & 4.38 & 2.82 && Pitcher base & 2.68 \\
     Wine glass & 3.85 & 2.87 && Bleach cleanser &3.56 \\
     Frying pan & 4.16 & 1.7 && Mug& 2.61 \\
     &&&& Power drill & 2.86 \\
     \bottomrule
    \end{tabular}
    \label{tab:AMT_study}
\end{table}

\section{Baselines Implementation Details}
\label{appen-baselines}
In Sec.~\ref{exp-wholegrasp} and Sec.~\ref{exp-motionfill}, we conduct several baseline experiments as comparisons with our WholeGrasp-VAE and MotionFill-VAE and also some ablation studies. In this section, we illustrate more implementation details about our baselines and ablation studies.
\subsection{Baselines to WholeGrasp-VAE}
\begin{figure}
    \centering
    \includegraphics[width=\linewidth]{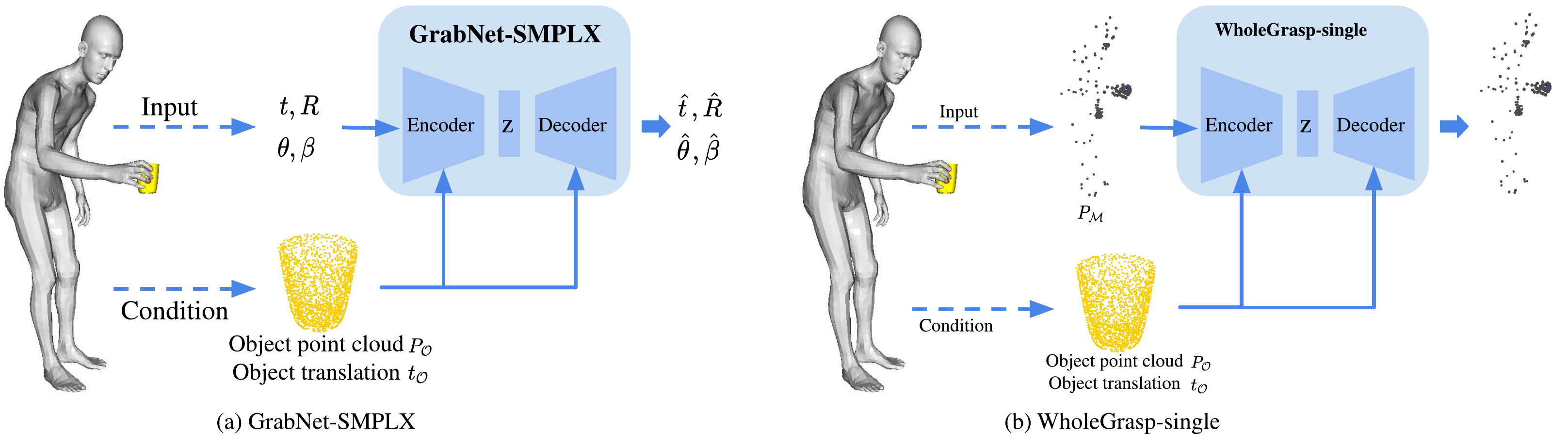}
    \caption{(a) The schematic architecture of GrabNet-SMPLX baseline. Similar as in GrabNet, we build a CVAE model to directly generate SMPL-X parameters; (b) The Schematic pipeline of the WholeGrasp-single baseline, which merely learns positions information of body markers.}
    \label{fig:baselines-wholegrasp}
\end{figure}

In Sec.~\ref{exp-wholegrasp}, we extend the GrabNet~\cite{GRAB:2020} to the whole-body grasping pose generation task (GrabNet-SMPLX) as a comparison with our WholeGrasp-VAE design, and we also study the effectiveness of the multi-task WholeGrasp-VAE by comparing with the single-task design (WholeGrasp-single). In the following, we provide detailed experimental setup of these two experiments.
\begin{itemize}
    \item \textbf{GrabNet-SMPLX.} GrabNet proposed to synthesize diverse hand grasps by directly learning the hand model parameters. And a similar idea can be extended to the full-body grasp synthesis by learning the compact SMPL-X body model parameters, which can include the body global configurations $t$ and $R$, the shape parameters $\beta$, and the full-body pose $\theta = [\theta_b, \theta_h, \theta_e]$. Fig.~\ref{fig:baselines-wholegrasp} (a) shows the schematic architecture of the GrabNet-SMPLX baseline. Different from the original GrabNet, instead of encoding the object shape using basic point set features~\cite{BPS}, we employ the PointNet++~\cite{qi2017pointnet++} in the GrabNet-SMPLX baseline, which is consistent with our WholeGrasp-VAE.
    \item \textbf{WholeGrasp-single.} As visualized in Fig.~\ref{fig:baselines-wholegrasp} (b), we build a single-task WholeGrasp-VAE, namely WholeGrasp-single, where we only learns the body markers positions. We employ the same multi-stage optimization algorithm as in GraspPose-Opt to fit SMPL-X body mesh to sampled markers. Recall that in our GraspPose-Opt, with the predicted body and object contact map, we design a contact loss accordingly (see Eq.~\ref{eq:opt-1-contact}) to refine the mutual contacts between the human body and object. However, due to the lack of contact map prediction in the single-task WholeGrasp-single, we cannot directly leverage this contact loss term to refine the hand pose. Instead of simply ignoring the contact refinement loss term in this test-time optimization step, we build a strong baseline by pre-defining a fixed hand contact pattern and designing a heuristic contact loss accordingly. Concretely, we firstly compute the average contact probability for each hand vertices over all the GRAB dataset, and we denote this hand contact prior probability as $C_\mathcal{H}$. Heuristically, we encourage those hand vertices that have a high prior contact probability (greater than 0.7) and also are closed enough to the object (less than 2cm) to contact with the object, and we formulate this heuristic contact loss baseline $E_{cont}^h$ as:
\begin{align}
    E_{cont}^h = \alpha_{cont}^h\sum_{h \in \mathcal{V}_{\bm{B}}^h}\mathbbm{1}(C_h > 0.7) \mathbbm{1}(d(h, \bm{O}) < 0.02) * C_hd(h, \bm{O}) \label{eq:contact_loss_heuristic}
\end{align}
where $\mathcal{V}_{\bm{B}}^h$ and $\bm{O}$ denote the hand vertices and object point cloud respectively, and $d(x, \mathcal{Y})=\min_{y\in\mathcal{Y}}||x-y||^2_2$.
Therefore, the overall optimization loss for the single-task WholeGrasp-single experiment is given by:
\begin{align}
    E_{opt}^{single}(\bm{\Theta}) = E_{fit}  + E_{colli}^o + E_{cont}^h + E_{cont}^g.
\end{align}
where $E_{fit}, E_{colli}^o, E_{cont}^g$ have the same formulations as in our GraspPose-Opt.

\end{itemize}

\subsection{Baselines to MotionFill-VAE}
In Sec.~\ref{exp-motionfill}, we compare our method with the convolution autoencoder network (CNN-AE) in~\cite{kaufmann2020convolutional}, LEMO~\cite{lemo2021}, and RouteNet and PoseNet from Wang \textit{et al.}~\cite{wang2021synthesizing}. For RouteNet and PoseNet, we remove the scene encoding branch from~\cite{wang2021synthesizing} and adopt the same route encoding branch and pose encoding branch architecture design. We use the same body representation as ours in all these experiments.

\section{Additional Results}
\label{appen-exp}
\subsection{Ablation Study on GraspPose-Opt optimization losses}
\begin{figure}
    \centering
    \includegraphics[width=1\linewidth]{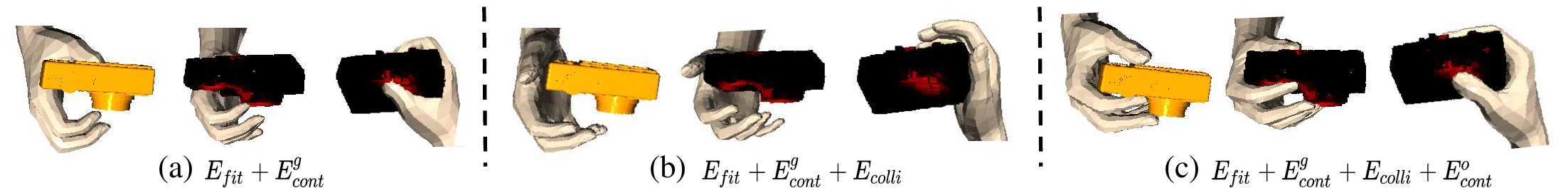}
    \caption{Visualization results of ablation study on the GraspPose-Opt optimization loss design in Table~\ref{tab:ablation-losses}. We present the optimized hand poses using different loss designs, and the red areas on the object surface indicate higher contact probability.}
    \label{fig:ablation_wholegrasp-opt_loss}
\end{figure}

In Sec.~\ref{exp-wholegrasp} and Table~\ref{tab:ablation-losses}, we have studied the effectiveness of our proposed GraspPose-Opt optimization loss design in Eq.~\ref{eq:opt-1} for optimizing human-object interactions. In Fig.~\ref{fig:ablation_wholegrasp-opt_loss}, we also present the visualization results of optimized hand poses using different loss designs to show the effects of our proposed loss terms. Since the hand pose can be highly sensitive to even tiny noises in markers positions, using only the basic markers fitting loss and foot ground contact loss, the recovered hand pose from markers can hardly interact with the object in a perceptually realistic way (see visualization result in Fig.~\ref{fig:ablation_wholegrasp-opt_loss} (a)). While the object collision loss $E_{colli}$ helps to mitigate the hand-object interpenetration issue (Fig.~\ref{fig:ablation_wholegrasp-opt_loss} (b)), the optimized hand does not grasp the object steadily. Using our mutual human-object contact loss $E_{cont}^o$, the object surface areas with higher contact probability attract hand vertices, and we can yield realistic and plausible hand-object interaction (see visualization result in Fig.~\ref{fig:ablation_wholegrasp-opt_loss} (c)).

\subsection{Additional Visualizations and Results on MotionFill-VAE}
\label{appen-exp-motioninfill}
\begin{figure}
    \centering
    \includegraphics[width=1\linewidth]{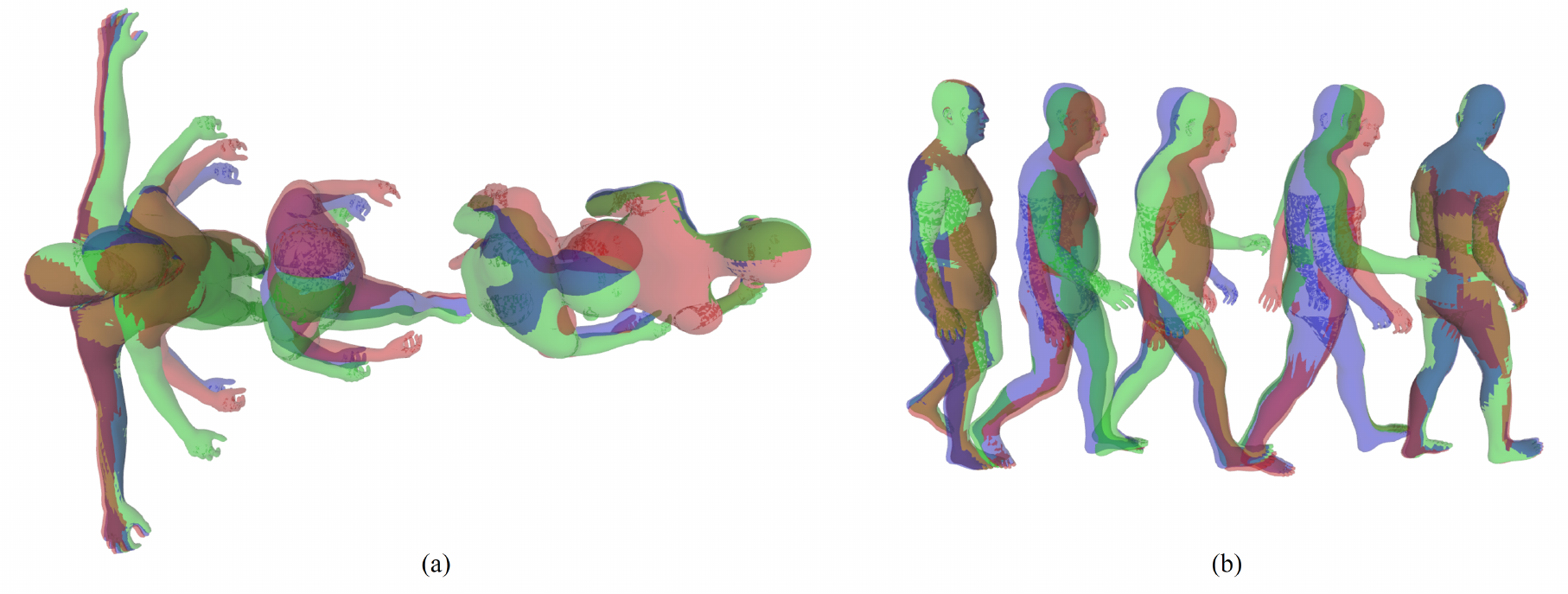}
    \caption{Visualization on generated diverse motion sequences. (a) Three motion sequences on three different trajectories generated by TrajFill-CVAE, respectively. (b) Three motion sequences generated on the same ground truth trajectory. Different colors (red, green, blue) represent different motion sequences in each sub-figure. The diverse intermediate frames show the stochasticity of our TrajFill-CVAE and LocalMotionFill-CVAE.}
    \label{fig:ablation_diverse}
\end{figure}

In Table~\ref{tab:exp-motionfillvae}, we have shown the quantitative results of our method compared with other state-of-the-art methods. In Fig.~\ref{fig:ablation_diverse}, we qualitatively present the diversity of the motions generated by our model which is finetuned on GRAB dataset~\cite{GRAB:2020}. The figure shows that our method can generate diverse trajectories as well as diverse local motions.

\begin{figure}
    \centering
    \includegraphics[width=1\linewidth]{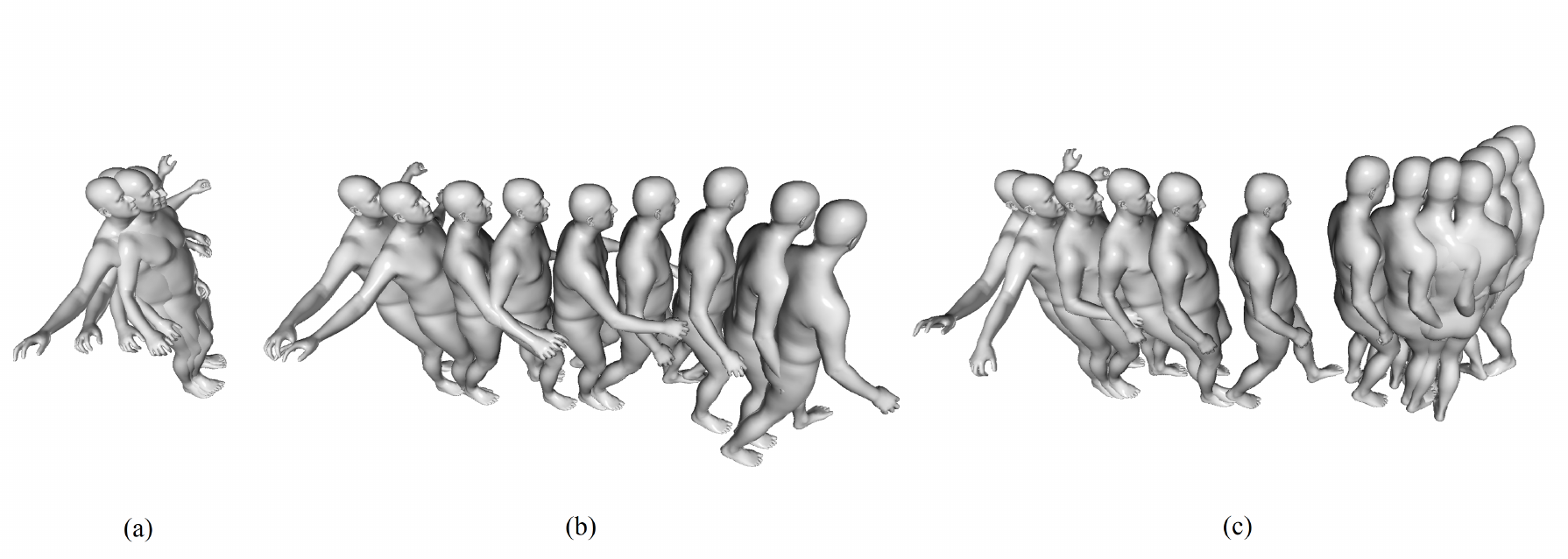}
    \caption{Visualization on generated motion sequences with different time lengths (2s, 4s, 6s). (a) 2-second motion sequences. (b) 4-second motion sequences. (c) 6-second motion sequences. We train these three MotionFill-VAE models using training data with different time lengths on AMASS dataset~\cite{AMASS:ICCV:2019}. The visualization results show that our motion infilling model can be easily generalized to different time horizons.}
    \label{fig:2s4s6s_motion}
\end{figure}

Limited by the short sequence length in the GRAB dataset, we only conduct the two-second motion infilling experiments with our MotionFill-VAE. Beyond generating two-second motion sequences given the starting pose and the ending pose, we show that our motion infilling model can be easily generalized to longer time lengths. Given the starting pose, ending pose, we train our MotionFill-VAE on AMASS dataset with 2s, 4s, 6s clips, respectively. In Fig.~\ref{fig:2s4s6s_motion}, we present the infilled motion sequences of 2 seconds, 4 seconds, and 6 seconds. The visualization results show that our motion infilling model is able to generate motions with different time lengths.

\section{Limitations}
\label{appen-limitations}
\begin{figure}
    \centering
    \includegraphics[width=0.8\linewidth]{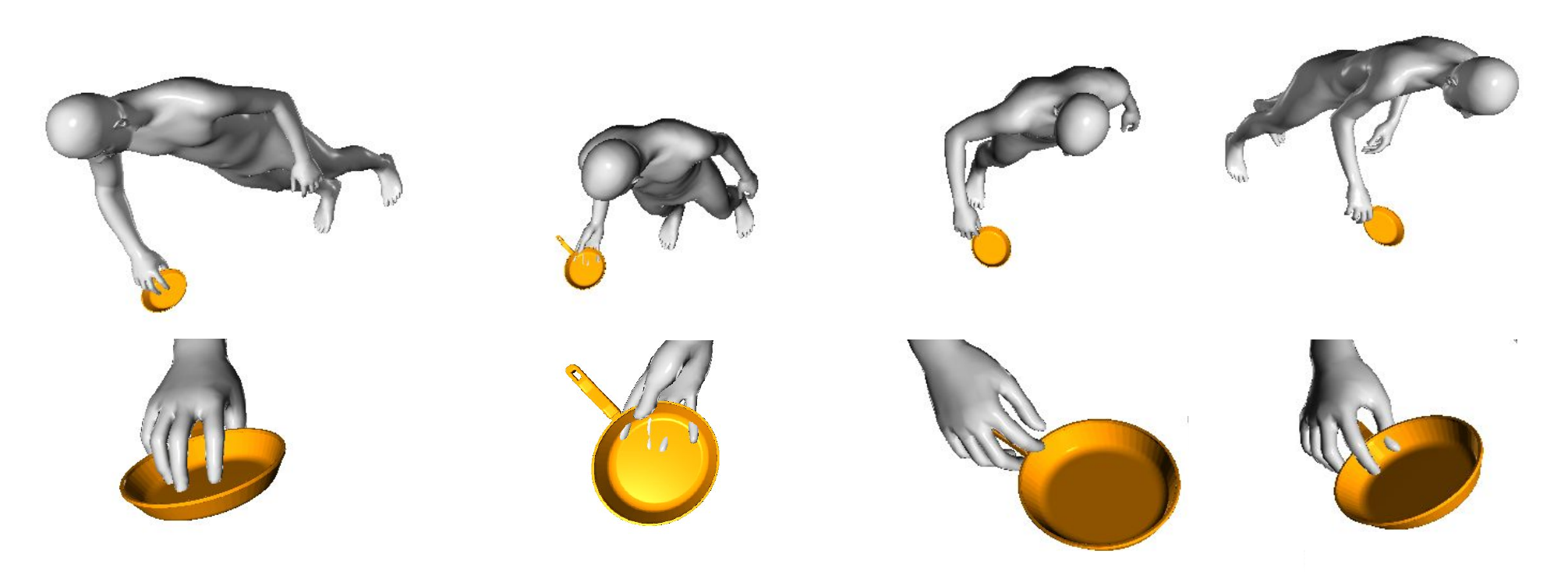}
    \caption{Grasping pose random samples for grasping the frying pan. Generating realistic grasping poses for frying pan pan is challenging. Although the generated humans are still in contact with the pan, they typically fails to grasp the handle of the pan, resulting in perceptually unrealistic results.}
    \label{fig:fryingpan-failure-case}
\end{figure}

Although our method can generate realistic grasping poses and grasping motions for most of the unseen objects in our test set, we observe some failure cases where the synthesized human fails to grasp the object in a realistic way. We have the similar observation as mentioned in GrabNet~\cite{GRAB:2020}, the frying pan is the most challenging object to grasp. As visualized in Fig.~\ref{fig:fryingpan-failure-case}, though the generated humans are in contact with the pan, they typically fail to grasp the pan handle, resulting in perceptually unrealistic results and low perceptual score in Table~\ref{tab:AMT_study}. 

\clearpage

\end{document}